\documentclass[12pt,a4paper]{article}

\pdfoutput=1

\RequirePackage[OT1]{fontenc}
\RequirePackage{amsmath}
\RequirePackage[round,comma]{natbib}
\RequirePackage[colorlinks,citecolor=blue,urlcolor=blue]{hyperref}
\RequirePackage{multicol}
\RequirePackage{enumitem}

\RequirePackage{multirow}
\usepackage{pdflscape}
\usepackage{array}
\newcolumntype{H}{>{\setbox0=\hbox\bgroup}c<{\egroup}@{}}

\RequirePackage{amssymb}
\RequirePackage{graphicx}

\RequirePackage{verbatim}
\usepackage{tabto}

\usepackage{adjustbox}
\usepackage{tabularx, booktabs}
\newcolumntype{Y}{>{\centering\arraybackslash}X}

\numberwithin{equation}{section}

\newcommand\argmax{\operatornamewithlimits{arg\,max}}

\newcommand{\red}[1]{{\color{red}{#1}}}
\newcommand{\green}[1]{{\color{green}{#1}}}
\newcommand{\blue}[1]{{\color{blue}{#1}}}

\def\bmx{{\mathbf x}}
\def\bmy{{\mathbf y}}
\def\bmz{{\mathbf z}}

\def\bmb{{\mathbf b}}

\def\bmX{{\mathbf X}}

\def\bmA{{\mathbf A}}

\def\bmC{{\mathbf C}}
\def\bmD{{\mathbf D}}
\def\bmH{{\mathbf H}}

\def\bmI{{\mathbf I}}

\def\bmSigma{{\mathbf{\Sigma}}}
\def\IR{\mathbb{R}}

\newcommand{\kNN}{\mbox{\emph{k-}NN}}
\def\LBL{\textsf{\textbf{---}}} 
\def\LBD{{\textbf{- -}}} 

\newtheorem{theorem}{Theorem}[section]  
\begin{document}

\title{\bigskip\bigskip\bigskip Classification with the pot-pot plot
}


\author{Oleksii Pokotylo         \and
        Karl Mosler
}



\maketitle

\begin{abstract}
We propose a procedure for supervised classification that is based on potential functions.
The potential of a class is defined as a kernel density estimate multiplied by the class's prior probability.
The method transforms the data to a potential-potential (pot-pot) plot, where each data point is mapped to a vector of potentials.
Separation of the classes, as well as classification of new data points, is performed on this plot. For this, either
the $\alpha$-procedure ($\alpha$-P) or $k$-nearest neighbors (\kNN) are employed. For data that are generated from continuous distributions, these classifiers prove to be
strongly Bayes-consistent.
The potentials depend on the kernel and its bandwidth used in the density estimate.
We investigate several variants of bandwidth selection, including joint and separate pre-scaling
and a bandwidth regression approach.
  The new method is applied to benchmark data from the literature, including simulated data sets as well as 50 sets of real data. It compares favorably to known classification methods such as LDA, QDA, max kernel density estimates, \kNN{}, and $DD$-plot classification using depth functions.
\end{abstract}

{\bf Key words:} Kernel density estimates, bandwidth choice, potential functions, 
$k$-nearest-neighbors classification, $\alpha$-procedure,
$DD$-plot, $DD\alpha$-classifier.

{\bf Mathematics Subject Classification:} 62H30, 62G07

\bigskip
\small{
\textbf{Acknowledgements}
We are grateful to Tatjana Lange and Pavlo Mozharovskyi for the active discussion of this paper. The work of Oleksii Pokotylo is supported by the Cologne Graduate School of Management, Economics and Social Sciences.
}

\noindent\rule[0.5ex]{\linewidth}{1pt}
\footnotesize{
O. Pokotylo \\
E-mail: pokotylo@wiso.uni-koeln.de

\smallskip
K. Mosler \\
E-mail: {mosler@statistik.uni-koeln.de}

\smallskip
Statistics and Econometrics\\
Universit{\"a}t zu K{\"o}ln\\
Albertus Magnus Platz,\\ 50923 K{\"o}ln, Germany
}

\section{Introduction}\label{sec:introduction}

Statistical classification procedures belong to the most useful and widely applied parts of statistical methodology. Problems of classification arise in many fields of application like economics, biology, medicine.
In these problems objects are considered that belong to $q\ge2$ classes. Each object has $d$ attributes and is represented by a point in $d$-space.
A finite number of objects is observed together with their class membership, forming $q$ training classes. Then, objects are observed whose membership is not known. The task of supervised classification consists in finding a rule by which any object with unknown membership is assigned to one of the classes.

A classical nonparametric approach to solve this task is by \textit{comparing kernel density estimates} (KDE); see e.g.\ \cite{Silverman86}.
The \textit{Bayes rule} indicates the class of an object $\bmx$ as $\argmax_j {(p_j f_j(\bmx))}$, where $p_j$ is the prior probability of class $j$ and $f_j$ its generating density.
In KDE classification, the density $f_j$ is replaced by a proper kernel estimate
$\hat f_{j}$ and a new object is assigned to a class $j$ at which its \textit{estimated potential},
\begin{equation}\label{eq_potential}
\hat\phi_j(\bmx)= p_j \hat f_j(\bmx)\,,
\end{equation}
is maximum. It is well known (e.g. \cite{DevroyeGL96})
that KDE is \textit{Bayes consistent}, that means, its expected error rate converges to the error rate of the Bayes rule for any generating densities.

As a practical procedure, KDE depends largely on the way by which the density estimates $\hat f_j$ and the priors $p_j$ are obtained. Many variants exist, differing in the choice of the multivariate kernel and, particularly, its bandwidth matrix. \cite{WandJ93} demonstrate that the choice of bandwidth parameters strongly influences the finite sample behavior of the KDE -classifier.
With higher-dimensional data, it is computationally infeasible to optimize a full bandwidth matrix. Instead, one has to restrict on rather few bandwidth parameters.

In this paper we modify the KDE approach by introducing a more flexible assignment rule in place of the maximum potential rule. We transform the data to a low-dimensional space, in which the classification is performed.
Each data point $\bmx$ is mapped to the vector\footnote{$\bmz^T$ denotes the transpose of $\bmz$.} $(\phi_{1}(\bmx), \dots, \phi_{q}(\bmx))^T$ in $\mathbb{R}^q_+$. The \textit{potential-potential plot}, shortly \textit{pot-pot plot}, consists of the transformed data of all $q$ training classes and the transforms of any possible new data to be classified.
With KDE, according to the maximum potential rule, this plot is separated into $q$ parts,
\begin{equation}\label{eq:maxpotrule}
\{\bmx\in\mathbb{R}^q_+ : j=\argmax_i {(\hat\phi_i(\bmx))}\}\,.
\end{equation}
If only two classes are considered,
the pot-pot plot is a subset of $\mathbb{R}^2_+$, where the coordinates correspond to potentials regarding the two classes.
Then, KDE corresponds to separating the classes by drawing the diagonal line in the pot-pot plot.

However, the pot-pot plot allows for more sophisticated classifiers. In representing the data, it reflects their proximity in terms of differences in potentials. To separate the training classes, we may apply any known classification rule
to their representatives in the pot-pot plot.
Such a separating approach, being not restricted to lines of equal potential, is able to provide better adapted classifiers.
Specifically, we propose to use either the \kNN-classifier or the $\alpha$-procedure to be used on the plot.
By the pot-pot plot procedure, the `curse of dimensionality' is bypassed, because - once the transformation and the separator have been established - any classification step is performed in $q$-dimensional space.

To construct a practical classifier, we first have to determine proper kernel estimates of the potential functions for each class. In doing this, the choice of a kernel, in particular of its bandwidth parameters, is a nontrivial task. It requires the analysis and comparison of many possibilities.
We evaluate them by means of the classification error they produce, using the following cross-validation procedure: Given a bandwidth matrix, one or more points are continuously excluded from the training data. The classifier is trained on the restricted data using the selected bandwidth parameter; then its performance is checked with the excluded points. The average portion of misclassified objects serves as an estimate of the classification error.
Notice that a kernel bandwidth minimizing this criterion may yield estimated densities that differ significantly from the actual generating densities of the classes.

Then we search for the optimal separation in $q$-dimensional space of the pot-pot plot.
This, in turn, allows us to keep the number of bandwidth parameters reasonably low, as well as the number of their values to be checked.

The principal achievements of this approach are:\vspace{-2mm}
\begin{itemize}
  \item The possibly high dimension of original data is reduced by the pot-pot transformation so that the classification can be done on a low-dimensional space, whose dimension equals the number of classes.
  \item The bias in kernel density estimates due to insufficiently adapted multivariate kernels is compensated by a flexible classifier on the pot-pot plot.
  \item In case of two classes, the proposed procedures are either always strongly consistent (if the final classifier is \kNN) or strongly consistent under a slight restriction (if the final classifier is the $\alpha$-classifier).
  \item The two procedures, as well as a variant that scales the classes separately, compare favourably with known procedures such as linear and quadratic discrimination and DD-classification based on different depths, particularly for a large choice of real data.
\end{itemize}
The paper is structured as follows.
Section \ref{sec:pf} presents density-based classifiers and the Kernel Discriminant Method.
Section \ref{sec:bandwidthSelection} treats the problem of selecting a kernel bandwidth.
 The pot-pot plot is discussed in Section \ref{sec:potpotplot}, and the consistency of the new procedures is established in Section \ref{sec:consistency}.
Section \ref{sec:scaling} presents a variant of pre-scaling the data, namely separate scaling.
Experiments with simulated as well as real data are reported in Section \ref{sec:experiments}.
Section \ref{sec:conclusion} concludes.

\section{Classification by maximum potential estimate}\label{sec:pf}

Comparing kernel estimates of the densities or potentials is a widely applied approach in classification.
Consider a data cloud $\bmX$ of points $\bmx_1,\dots,\bmx_n\in \mathbb{R}^d$ and assume that the cloud is generated as an independent sample from some probability density $f$.
The potential of a given point $\bmx\in\mathbb{R}^d$ regarding the data cloud $\bmX$ is estimated by a kernel estimator.

Let $K_{\bmH}(\bmx,\bmx_{i}) =|\det \bmH|^{-1/2} K\left(\bmH^{-1/2} \, (\bmx - \bmx_{i})\right)$,
$\bmH$ be a symmetric and positive definite bandwidth matrix, and
$K: \mathbb{R}^d\to [0,\infty[$ be a spherical probability density function, $K(\bmz)= r(\bmz^T\bmz)$, with $r:[0,\infty[ \to [0,\infty[$ non-increasing and bounded.
Then
\begin{eqnarray} \label{eq:f_sumK1}
\hat f_\bmX(\bmx) &=& \frac{1}{n} \sum_{i=1}^{n}{K_{\bmH}(\bmx,\bmx_{i})}\\
&=& \frac 1n \sum_{i=1}^{n} |\det \bmH|^{-\frac 12} K\bigl(\bmH^{-\frac 12}(\bmx-\bmx_i)\bigr)\nonumber
\end{eqnarray}
is a kernel estimator of the density of $\bmx$ with respect to $\bmX$.
In particular, the Gaussian function
$K(\bmz)=(2\pi)^{-d/2}\exp\left(-\frac{1}{2}\bmz^T\bmz \right)$ will be employed below.


Let $\bmH_\bmX$ be an affine invariant estimate of the dispersion of $\bmX$, that is,
\begin{equation}\label{eq:affeqH}
     \bmH_{\bmA\bmX+ \bmb} = \bmA \bmH_\bmX \bmA^T \quad \text{for any $\bmA$ of full rank and $\bmb\in \IR^d$}\,.
\end{equation}
Then $\bmH_{\bmA\bmX + \bmb}^{-\frac 12}=(\bmA \bmH_\bmX \bmA^T)^{-\frac 12}= (\bmA \bmH_\bmX^{\frac 12})^{-1} = \bmH_\bmX^{-\frac 12}\bmA^{-1}$ and
\begin{eqnarray*}
\phi_{\bmA\bmX+\bmb}(\bmA\bmy+\bmb)
&=& \frac 1n \sum_{i=1}^{n} |\det \bmA|^{-1} |\det \bmH_\bmX|^{-\frac 12} K\bigl(\bmH_\bmX^{-\frac 12}\bmA^{-1}(\bmA\bmy-\bmA\bmx_i)\bigr) \\
 &=& |\det \bmA|^{-1} \phi_\bmX(\bmy)
\end{eqnarray*}
Hence, in this case, the potential is affine invariant, besides a constant factor $|\det \bmA|^{-1}$.

\newpage  
\textbf{Examples}\begin{itemize}[topsep=0pt]
                   \item If $\bmH_\bmX= h^2 \hat \Sigma_\bmX$, (\ref{eq:affeqH}) is satisfied; the potential is affine invariant (besides a factor).
                   \item If $\bmH_\bmX= h^2 \bmI$ and $\bmA$ is orthogonal, we obtain $\bmH_{\bmA\bmX+ \bmb}= h^2 \bmI=h^2 \bmA\bmA^T=\bmA \bmH_\bmX \bmA^T$, hence (\ref{eq:affeqH}); the potential is \textit{orthogonal invariant}.
                   \item If $\bmH_\bmX=h^2 diag(\hat\sigma_1^2,\dots,\hat\sigma_d^2)$ and $\bmA= diag(a_1,\dots,a_d)$, then $\bmH_{\bmA\bmX+ \bmb}=h^2 diag(a_1^2 \hat\sigma_1^2,\dots,a_d^2 \hat\sigma_d^2) = \bmA \bmH_\bmX \bmA^T$; the potential is \textit{invariant regarding componentwise scaling}.
                 \end{itemize}
The selection of the bandwidth matrices $\bmH$ is further discussed in Section \ref{sec:bandwidthSelection}.

\begin{figure}[h]
        \centering
        \includegraphics[width=1\textwidth,natwidth=3102,natheight=957]{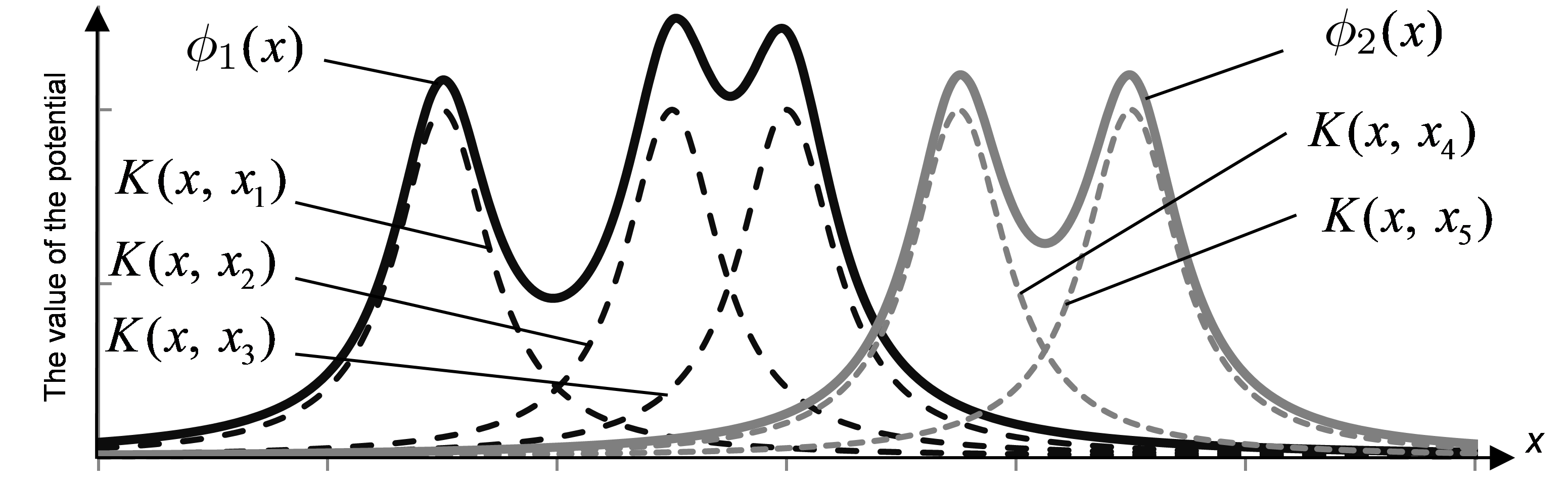}
        \caption{Potentials of two classes, $\{\bmx_1, \bmx_2,\bmx_3\}$ and $\{\bmx_{4}, \bmx_{5}\}$.
        }
        \label{f:potfunc}
    \end{figure}

Now, consider a classification problem with $q$ training classes $\bmX_1,\dots, \bmX_q$, generated by densities $f_1,\dots, f_q$, respectively.
Let the class $\bmX_j$ consist of points $\bmx_{j1},\dots,\bmx_{jn_j}$.
The potential of a point $\bmx$ with respect to $\bmX_j$ is estimated by
\begin{equation} \label{eq:f_sumKj}
\hat\phi_j(\bmx) = p_j \hat f_j(\bmx) = \frac{1}{n} \sum_{i=1}^{n_j}{K_{\bmH_j}(\bmx,\bmx_{ji})},
\end{equation}
$j=1,\dots, q$. Figure \ref{f:potfunc} exhibits the potentials of two classes $A_1 = \{\bmx_1, \bmx_2,\bmx_3\}$ and $A_2 = \{\bmx_{4}, \bmx_{5}\}$.

The Bayes rule yields the class index of an object $\bmx$ as $\argmax_j {(p_j f_j(\bmx))}$, where $p_j$ is the prior probability of class $j$.
The \textit{potential discriminant rule} mimics the Bayes rule:
Estimating the prior probabilities by $p_j = n_j/\sum_{k=1}^q n_k$ it yields
\begin{equation}\label{eq_kerneldiscriminant}
\argmax_j {(p_j \hat f_j(\bmx))} = \argmax_j {\sum_{i=1}^{n_j}{K_{\bmH_j}(\bmx,\bmx_{ji})}}.
\end{equation}

\section{Multivariate bandwidth}\label{sec:bandwidthSelection}

\begin{figure}[h]
        \centering
        \includegraphics[keepaspectratio=true,width=.9\textwidth]{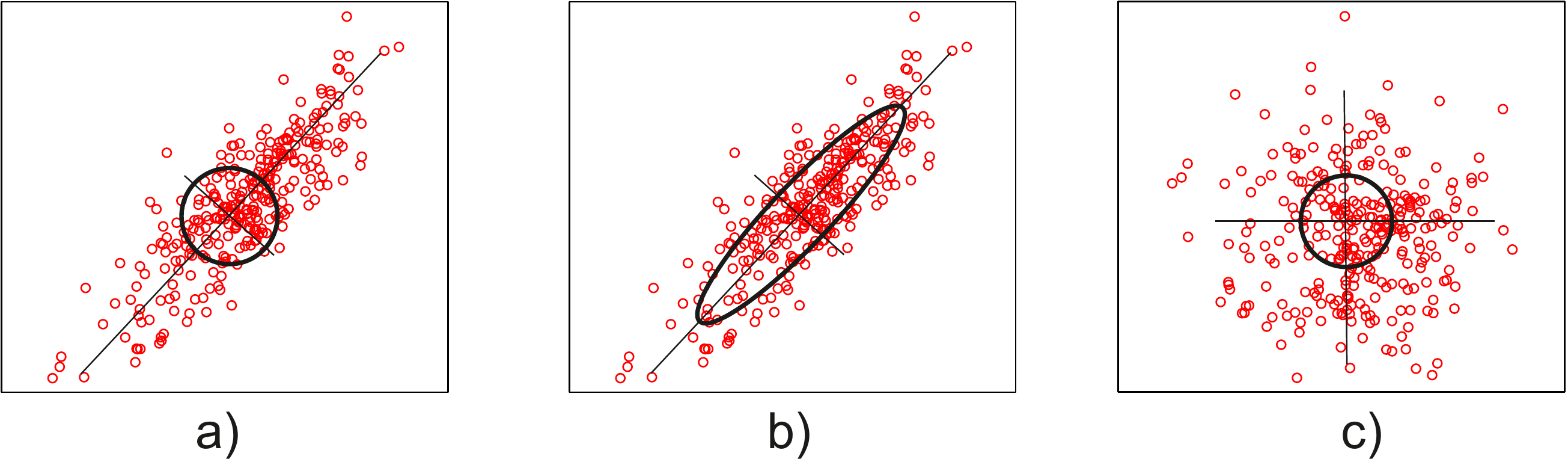}
        \caption{Different ways of fitting a kernel to the data:
a) applying a spherical kernel to the original data;
b) applying an elliptical kernel to the original data;
c) applying a spherical kernel to the transformed data.}
        \label{f:sphering_data}
    \end{figure}

The kernel bandwidth controls the range on which the potentials change with a new observation.
\citet{AizermanBR70}, among others, use a kernel with bandwidth matrix $\bmH_j=h^2 \bmI$ for both classes and apply this kernel to the data as given.
This means that the kernels are spherical and treat the neighborhood of each point equally in all directions.
However, the distributions of the data are often not close to spherical,
thus with a single-parameter spherical kernel the estimated potential differs from the real one more in some directions than in the others (Figure \ref{f:sphering_data}.a).
In order to fit the kernel to the data a proper bandwidth matrix $\bmH$ is selected (Figure \ref{f:sphering_data}.b).
This matrix $\bmH$ can be decomposed into two parts, one of which follows the shape of the data, and the other the width of the kernel. Then the first part may be used to transform the data, while the second is employed as a parameter of the kernel and tuned to achieve the best separation
(Figure \ref{f:sphering_data}.b,c).

\cite{WandJ93} distinguish several types of bandwidth matrices to be used in (\ref{eq:f_sumK1}) and (\ref{eq:f_sumKj}):
    a spherically symmetric kernel bandwidth matrix $\bmH_1 = h^2\bmI$ with one parameter; a matrix with $d$ parameters $\bmH_2 = diag(h_1^2,h_2^2, ...,h_d^2)$, yielding kernels that are elliptical along the coordinate axes; and an unrestricted symmetric and positive definite matrix $\bmH_3$ having $\frac{d(d+1)}{2}$ parameters, that produces elliptical kernels with an arbitrary orientation.
With more parameters the kernels are more flexible, but require more costly tuning procedures.
The data may also be transformed beforehand using their mean $\bar \bmx$ and either the marginal variances $\hat\sigma^2_{i}$ or the full empirical covariance matrix $\hat\bmSigma$.
These approaches are referred to as \textit{scaling} and \textit{sphering}.
They employ the matrices
 $\bmC_2 = h^2\hat\bmD$ and $\bmC_3 = h^2\hat\bmSigma$,  respectively,
 where $\hat\bmD= diag(\hat\sigma^2_{1},\dots, \hat\sigma^2_{d})$.
The matrix $\bmC_2$ is a special case of $\bmH_2$, and the matrix $\bmC_3$ is a special case of $\bmH_3$.
Each has only one tuning parameter $h^2$ and thus the same tuning complexity as $\bmH_1$, but fits the data much better.
    Clearly, the  bandwidth matrix $\bmC_3 = h^2{\hat\bmSigma}$ is equivalent to the bandwidth matrix  $\bmH_1 = h^2\bmI$ applied to the pre-scaled data $\bmx^\prime = {\hat\bmSigma}^{-\frac{1}{2}} (\bmx-\bar\bmx)$.

\cite{WandJ93} show by experiments  that sphering with one tuning parameter $h^2$ shows poor results compared to the use of $\bmH_2$ or $\bmH_3$ matrices. \cite{Duong07} suggests to employ at least a diagonal bandwidth matrix $\bmH^* = diag(h_1^2,h_2^2, ...,h_n^2)$ together with a scaling transformation,
$\bmH = \hat\bmSigma^{1/2} \bmH^* \hat\bmSigma^{1/2}$.
But, even in this simplified procedure the complexity of the parameter tuning grows exponentially with the number of attributes, that is the dimension of the data.

In univariate density estimation the bandwidth $h^2$ is often chosen by a rule of thumb \citep{Silverman86},
\begin{equation}\label{eq_univariaterule}
    h^2=\left(\frac 4{d+2}\right)^{2/(d+4)} n^{-2/(d+4)} \hat\sigma^2,
\end{equation}
which is based on an approximative normality assumption. As the first factor in (\ref{eq_univariaterule})
is almost equal to one, the rule is further simplified to Scott's rule \citep{Scott92},  $h^2 = n^{-2/(d+4)}\hat\sigma^2.$
In multivariate estimation, disregarding the covariance of the data, these rules may be separately applied to each component $h_i^2$ \citep{HardleMSW04}, which is of matrix type $\bmH_2$.
If the covariance structure is not negligible, the generalized Scott's rule may be used, having matrix
\begin{equation}\label{eq_genScott}
  \bmH_s=n^{-2/(d+4)}\hat\bmSigma\,.
\end{equation}
Observe that the matrix $\bmH_s$ is of type $ \bmC_3$.
Equivalently, after sphering the data with $\hat\bmSigma$, a bandwidth matrix of type $\bmH_1$ is applied with $h^2=n^{-2/(d+4)}$.

 Here we propose procedures that employ one-parameter bandwidths combined with sphering transformations of the data. While this yields rather rough density estimates, the imprecision of the potentials is counterbalanced by a sophisticated non-linear classification procedure on the pot-pot plot. 
The parameters tuning procedure works as follows:
 The bandwidth parameter is systematically varied over some range, and a value is selected that gives smallest classification error.

\section{Pot-pot plot classification} \label{sec:potpotplot}

In KDE classification a new object is assigned to the class that grants it the largest potential.
A pot-pot plot allows for more sophisticated solutions.
By this plot, the original $d$-dimensional data is transformed to $q$-dimensional objects.
Thus the `curse of dimensionality' is bypassed and the classification is performed in $q$-dimensional space.

E.g. for $q=2$, denote the two training classes as $\bmX_1=\{\bmx_{1}, \dots, \bmx_{n}\}$ and $\bmX_2=\{\bmx_{n+1}, \dots, \bmx_{n+m}\}$.
Each observed item corresponds to a point in $\mathbb{R}^d$.
The pot-pot plot $Z$ consists of the potential values of all data w.r.t.\ the two classes.
\[Z=\{\bmz_i=(z_{i1},z_{i2}) : z_{i1}=\phi_1(\bmx_{i}),z_{i2}=\phi_2(\bmx_{i}),i=1,...,n+m\}\,.\]
Obviously, the maximum-potential rule results in a diagonal line separating the classes in the pot-pot plot.

However, any classifier can be used instead for a more subtle separation of the classes in the pot-pot plot.
Special approaches to separate the data in the pot-pot plot are: using $k$-nearest neighbors (\kNN) or
linear discriminant analysis (LDA), regressing a polynomial line, or employing the \textit{$\alpha$-procedure}.
The $\alpha$-procedure is a fast heuristic that yields a polynomial separator; see \cite{LangeMM14a}.
Besides \kNN, which classifies directly to $q\ge 2$ classes in the pot-pot plot, the other procedures classify to $q=2$ classes only. If $q>2$, several binary classifications have to be performed, either $q$ `one against all' or $q(q-1)/2$ `one against one', and be aggregated by a proper majority rule.

Recall that our choice of the kernel needs a cross-validation of the single bandwidth parameter $h$.
For each particular pot-pot plot an optimal separation is found by selecting the appropriate number of neighbors for the \kNN{}-classifier, or the degree of the $\alpha$-classifier.
For the $\alpha$-classifier a selection is performed in the constructed pot-pot plot, by dividing its points into several subsets, sequentially excluding one of them, training the pot-pot plot classifier using the others and estimating the classification error in the excluded subset.
For the \kNN{}-classifier an optimization procedure is used that calculates the distances from each point to the others, sorts the distances and estimates the classification error for each value of $k$.
The flexibility of the final classifier compensates for the relative rigidity of the kernel choice.

Our procedure bears an analogy to \textit{$DD$-classification}, as it was introduced by \cite{LiCAL12}. There, for each pair of classes,
the original data are mapped to a two-dimensional depth-depth (DD) plot and classified there.
A function $\bmx \mapsto D^d(\bmx|\bmX)$ is used that indicates how central a point $\bmx$ is situated in a set $\bmX$ of data or, more general, in the probability distribution of a random vector $\bmX$ in $\mathbb{R}^d$.
The upper level sets of $D^d(\cdot|\bmX)$ are regarded as `central regions' of the distribution.
$D^d(\cdot|\bmX)$ is called a \emph{depth function} if its \textit{level sets} are
\begin{itemize}[topsep=0pt]
  \item \textit{closed} and \textit{bounded},
  \item \textit{affine equivariant}, that is, if $\bmX$ is transformed to $\bmA\bmX+\bmb$ with some regular matrix $\bmA\in \IR^{d\times d}$ and $\bmb \in \IR^d$, then the level sets are transformed in the same way.
\end{itemize}
Clearly, a depth function is \textit{affine invariant}; $D^d(\bmx|\bmX)$ does not change if both $\bmx$ and $\bmX$ are subjected to the same affine transformation.   For surveys on depth functions and their properties, see e.g.\ \cite{ZuoS00}, \cite{Serfling06}, and \cite{Mosler13}.

 More generally, in \textit{$DD$-classification}, the depth of all data points is determined with respect to each of the $q$ classes, and a data point is represented in a $q$-variate $DD$-plot by the vector of its depths. Classification is done on the $DD$-plot, where different separators can be employed.
In \cite{LiCAL12} a polynomial line is constructed, while \cite{LangeMM14a} use the $\alpha$-procedure and \cite{Vencalek13} suggests to apply $\kNN$ in the depth space.
Similar to the polynomial separator of \cite{LiCAL12}, the $\alpha$-procedure results in a polynomial separation, but is much faster and produces more stable results. Therefore we focus on the $\alpha$-classifier in this paper.

Note that \cite{FraimanM99} mention a density estimate $\hat f_\bmX(\bmx)$ as a \textit{`likelihood depth'}. Of course this `depth' does not satisfy the usual depth postulates.
Principally, a depth relates the data to a `center' or `median', where it is maximal; a `local depth' does the same regarding several centers. \cite{PaindaveineVB13} provide a local depth concept that bears a connection with local centers.
Different from this, a density estimate measures at a given point how much mass is located around it; it is of a local nature, but not related to any local centers. This fundamental difference has consequences in the use of these notions as descriptive tools as well as in their statistical properties, e.g. regarding consistency of the resulting classifiers; see also \cite{PaindaveineVB15}.

Maximum-depth classification with the `likelihood depth' (being weighted with prior probabilities) is the same as KDE. \cite{CuevasFF07} propose an extension of this notion to functional data, the h-depth that is calculated as $\hat D^d(\bmx) = \frac{1}{n} \sum_{i=1}^{n}{K\left(\frac{m(\bmx,\bmx_{i})}{h}\right)}$, where $m$ is a distance. The h-depth is used in \cite{CuestaAlbertosFBOdlF15}, among several genuine depth approaches, in a generalized DD-plot to classify functional data.
 However, the DD$^G$ classifier with h-depth applies equal spherical kernels to both classes, with the same parameter $h$.
The authors also do not discuss about the selection of $h$, while \cite{CuevasFF07} proposed keeping it constant for the functional setup.
Our contribution differs in many respects from the latter one:
(1) We use Gaussian kernels with data dependent covariance structure and optimize their bandwidth parameters. (2) The kernel is selected either simultaneously or separately for each class. (3) When $q=2$, in case of separate sphering, a regression between the two bandwidths is proposed, that allows to restrict the optimization to just one bandwidth parameter (see sec. \ref{sec:scaling}).
(4) Strong consistency of the procedure is demonstrated (see the next Section). (5) The procedure is compared with known classification procedures on a large number of real data sets (see sec. \ref{sec:experiments}).

\section{Bayes consistency}\label{sec:consistency}

We advocate the pot-pot procedure as a data-analytic tool to classify data of unknown origin, generally being non-normal, asymmetric and multi-modal.
Nevertheless, it is of more than theoretical interest, how the pot-pot procedure behaves when the sample size goes to infinity.
Regarded as a statistical regression approach, Bayes consistency is a desirable property of the procedure.

We consider the case $q=2$. The data are seen as realizations of a random vector $(\bmX, Y)\in \mathbb{R}^d\times \{1,2\}$, that has probability distribution $P$.
A classifier is any function $g: \mathbb{R}^d\to \{1,2\}$.
Notate $p_j(\bmx)=P(Y=j|\bmX= \bmx)$.
The \textit{Bayes classifier} $g^*$ is given by $g^*(\bmx)= 2$ if $p_2(\bmx) > p_1(\bmx)$ and $g^*(\bmx)=1$ otherwise.
Its probability of misclassification, $P(g^*(\bmX) \not= Y)$ is the best achievable risk, which is named the \textit{Bayes risk}.

We assume that the distributions of $(\bmX, 1)$ and $(\bmX, 2)$ are continuous.
Let the potentials be estimated by a continuous regular kernel (see Definition 10.1 in \cite{DevroyeGL96}), like a Gaussian kernel, and let $(h_n)$ be a
sequence of univariate bandwidths satisfying
\begin{equation}\label{suffconsistencykernel}
  h_n\to 0 \quad \text{and} \quad nh_n\to \infty\,.
\end{equation}
It is well-known (see Theorem 10.1 in \cite{DevroyeGL96}) that then the maximum potential rule is \textit{strongly Bayes-consistent}, that is,
its error probability almost surely approaches the Bayes risk for any distribution of the data.
The question remains, whether the proposed procedures operating on the pot-pot plot attain this risk asymptotically.

We present two theorems about the Bayes consistency of the two variants of the pot-pot classifier.
\begin{theorem}[$k$-$NN$]
Assume that  $(\bmX, 1)$ and $(\bmX, 2)$ have continuous distributions.
Then the pot-pot procedure is strongly Bayes consistent if the separation on the pot-pot plot is performed by
$k$-nearest neighbor classification with $k_n\to \infty$ and $k_n/n\to 0$.
\end{theorem}

\textbf{Proof:} \ Let us first define a sequence of pot-pot classifiers that satisfies (\ref{suffconsistencykernel}).
We start with two training classes of sizes $n_1^*$ and $n_2^*$, set $n^*=n_1^*+n_2^*$, and determine a proper bandwidth $h^*$ by cross-validation as described above. Then, let $n_1\to \infty$ and $n_2\to \infty$, $n=n_1+n_2$. For $n>n^*$ we restrict the search for $h_n$ to the interval
\[\left[h^*\cdot n^{\frac{\epsilon-1}d},   h^* \cdot n^{\frac{\delta-1}d}\right],
\]
with some $0< \epsilon\le \delta< 1$. It follows that $h_n\to 0$ and $nh_n\to\infty$ as $n$ goes to infinity, which yields the a.s. strong Bayes consistency of the maximum potential rule.
The maximum potential rule corresponds to the diagonal of the pot-pot plot. This separator almost surely asymptotically attains the Bayes risk.

We have still to demonstrate that the $k$-nearest neighbor procedure applied to the transformed data on the pot-pot plot yields the same asymptotic risk.
Under $k_n\to \infty$ and $k_n/n\to 0$, the $k$-$NN$ procedure on the pot-pot plot is strongly Bayes consistent
if either $\phi_1(\bmX)$ or $\phi_2(\bmX)$ is continuously distributed in $\mathbb{R}^2$; see Theorem 11.1 and page 190 in  \cite{DevroyeGL96}.
But the latter follows from the continuity of the regular kernel.
Obviously, the Bayes risk of classifying the transformed data is the same as that of classifying the original data.
It follows that for any distribution of the original data the pot-pot procedure achieves the Bayes risk almost surely asymptotically. \hfill $\Box$

\begin{theorem}[$\alpha$-procedure]
Assume that  $(\bmX, 1)$ and $(\bmX, 2)$ have continuous distributions and that
\begin{equation}\label{eq:probdiag}
P(p_1(\bmX)=p_2(\bmX)) =0\,.
\end{equation}
Then the pot-pot procedure is strongly Bayes consistent if the separation on the pot-pot plot is performed by the $\alpha$-procedure.
\end{theorem}

\textbf{Proof:} \ As in the preceding proof,
the maximum potential rule corresponds to the diagonal of the pot-pot plot, and this separator almost surely asymptotically attains the Bayes risk.
Consider the sign of the difference between the two (estimated) potentials. If the sample size goes to infinity the number of 'wrong' signs goes to zero. By assumption (\ref{eq:probdiag}) also the number of ties (corresponding to points on the diagonal) goes to zero.
By definition, the $\alpha$-procedure in its first step considers all pairs of features, the original $z_1$ and $z_2$ and possibly polynomials of them up to some pre-given degree. Then for each pair a separating line is determined that minimizes the empirical risk; see \cite{LangeMM14a}.
Thus, once the differences of the potentials have the correct sign, the $\alpha$-procedure will produce the diagonal of the $(z_1,z_2)$-plane (or a line that separates the same points) in its very first step.\hfill $\Box$

Compared to these results, the Bayes consistency of depth-depth (DD) plot procedures is rather limited. It has been established only if both classes follow unimodal elliptically symmetric distributions; see \cite{LiCAL12} and \cite{LangeMM14a}.
The reason is that depth functions fit primarily to unimodal distributions.
Under unimodal ellipticity, since a depth function is affine invariant, its level sets are ellipsoids that correspond to density level sets, and the depth is a monotone function of the density.

\begin{figure}[h]
    \centering
    \includegraphics[keepaspectratio=true,width = 0.3\textwidth, page = 2]{ddplot_avsd.pdf}
    ~~~~~~~~~~~
    \includegraphics[keepaspectratio=true,width = 0.3\textwidth, page = 1]{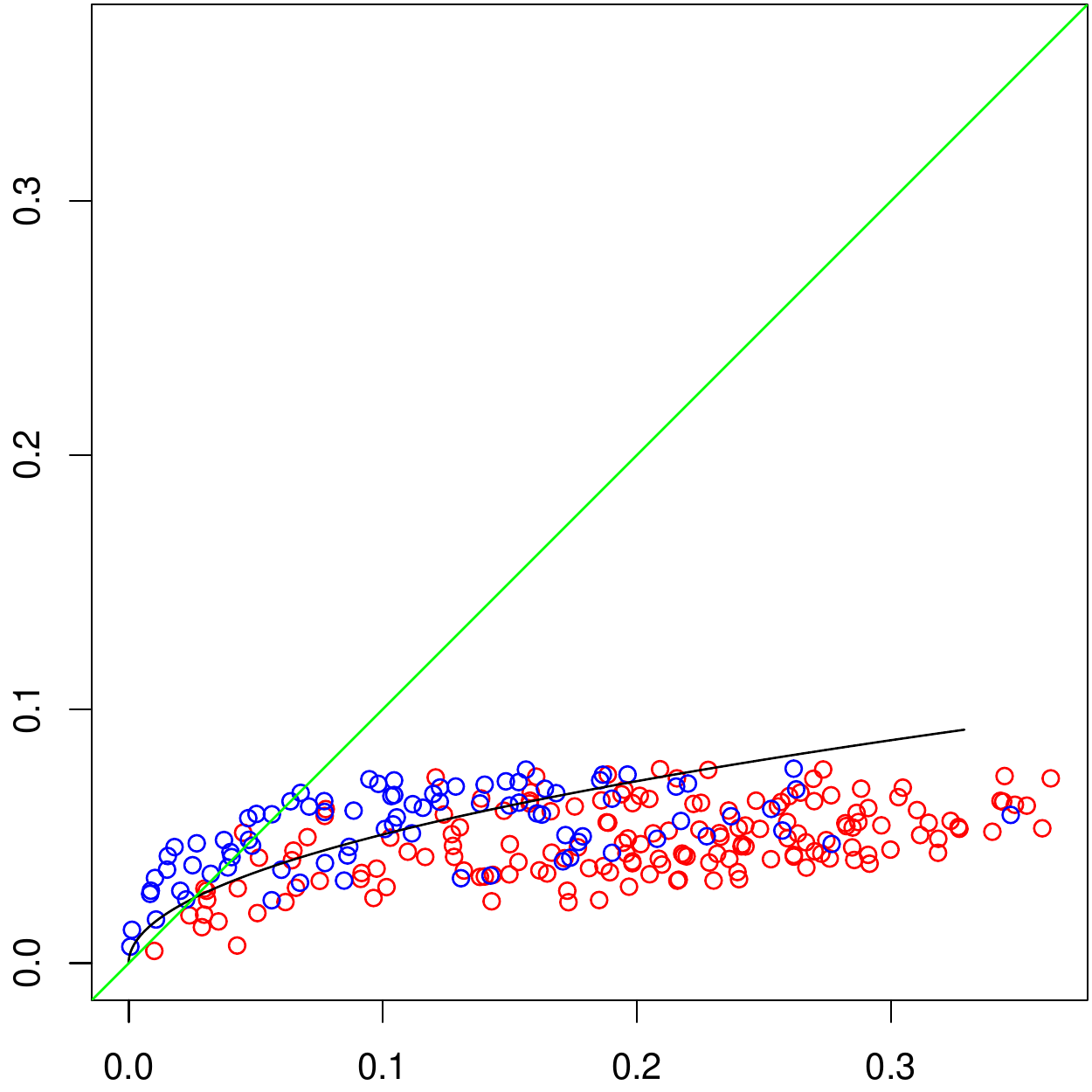}
    \caption{
        Examples of $\alpha$-separation in the pot-pot plot. In the left panel (data set `tennis') the $\alpha$-classifier coincides with the diagonal, while in the right panel (data set `baby') it provides a completely different separation. Bandwidth parameters are selected according to best performance of the $\alpha$-classifier.
    }
    \label{f:ddplot_avsd}
\end{figure}

Note that the distributions of the two training classes, after having been transformed by the `sphering transformation', can be still far away from being spherical. This happens particularly often with real data. It is well known that with a single parameter the multidimensional potentials are often poorly estimated by their kernel estimates. Then the best separator may differ considerably from the diagonal of the pot-pot plot as our results will demonstrate, see Figure \ref{f:ddplot_avsd}.
The final classification on the plot compensates the insufficient estimate by searching for the best separator on the plot.

\newpage 
\section{Scaling the data} \label{sec:scaling}

In Section \ref{sec:bandwidthSelection} we have shown that it is convenient to divide the bandwidth matrix into two parts, one of which is used to scale the data, and the other one to tune the width of a spherical kernel.
The two classes may be scaled jointly or separately, before proper bandwidth parameters are tuned.
Note that \cite{AizermanBR70} do not scale the data and use the same kernel for both classes.

KDE naturally estimates the densities individually for each class and tunes the bandwidth matrices separately. On the other hand, SVM scales the classes jointly \citep{ChangL11}, either dividing each attribute by its standard deviation, or scaling it to $[0;1]$.

\begin{figure}[h]
    \centering
    \includegraphics[keepaspectratio=true,width = 0.95\textwidth, page = 1]{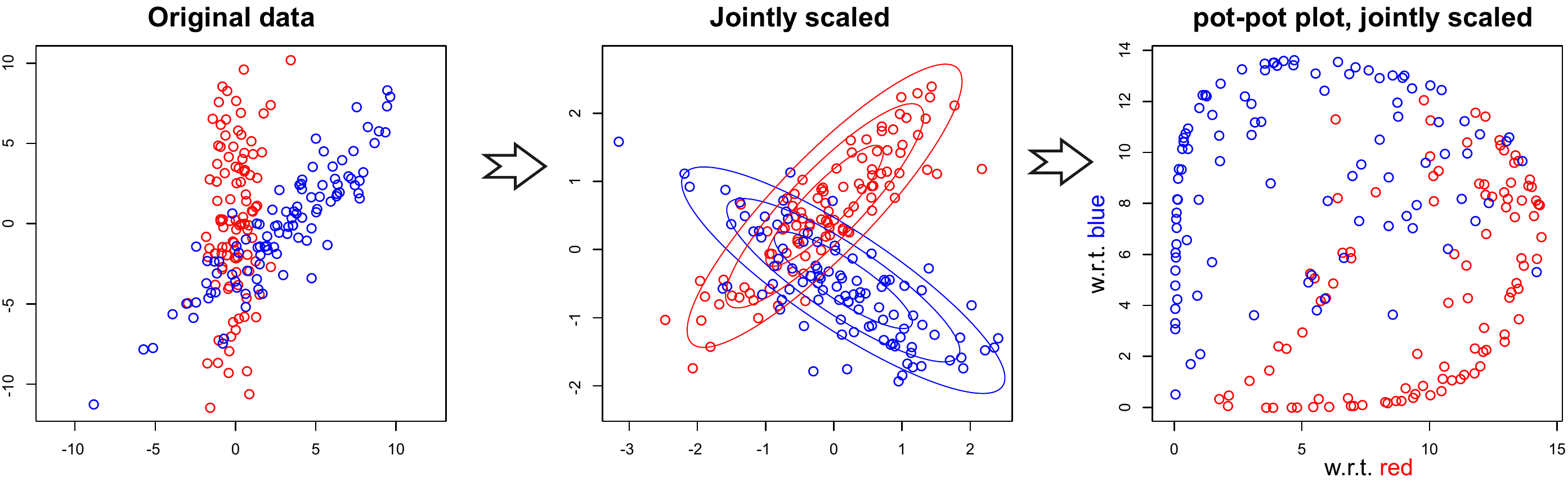}
    \caption{
        The data is jointly scaled.
        The plots show the original data,
            the scaled data with their lines of equal potential,
            and the corresponding \mbox{pot-pot plot}.
            }
    \label{f:scaling_joint}
\end{figure}

In what follows we consider two approaches: joint and separate scaling.
With \textit{joint scaling} the data is sphered using a proper estimate $\hat\bmSigma$ of the covariance matrix of the merged classes; then a spherical kernel of type $\bmH_1$ is applied.
This results in potentials
\[\phi_j(\bmx)= p_j \hat f_j(\bmx) = p_j \frac 1{n_j} \sum_{i=1}^{n_j}{K_{h^2 \hat \bmSigma}(\bmx-\bmx_{ji})}\,,
\]
{where an estimate $\hat\bmSigma$ of the covariance matrix has to be calculated and one scalar parameter $h^2$ has to be tuned.
(Note that $\hat\bmSigma$ is not necessarily the empirical covariance matrix; in particular, some more robust estimate may be used.) }
The scaling procedure and the obtained \mbox{pot-pot plot} are illustrated in Fig. \ref{f:scaling_joint}.
Obviously, as the classes differ, the result of joint scaling
is far away from being spherical and the spherical kernel does not fit the two distributions well.
However, these kernels work well when the classes' overlap is small; in this case the separation is no big task.

\begin{figure}[h]
    \centering
    \includegraphics[keepaspectratio=true,width = 0.95\textwidth, page = 4]{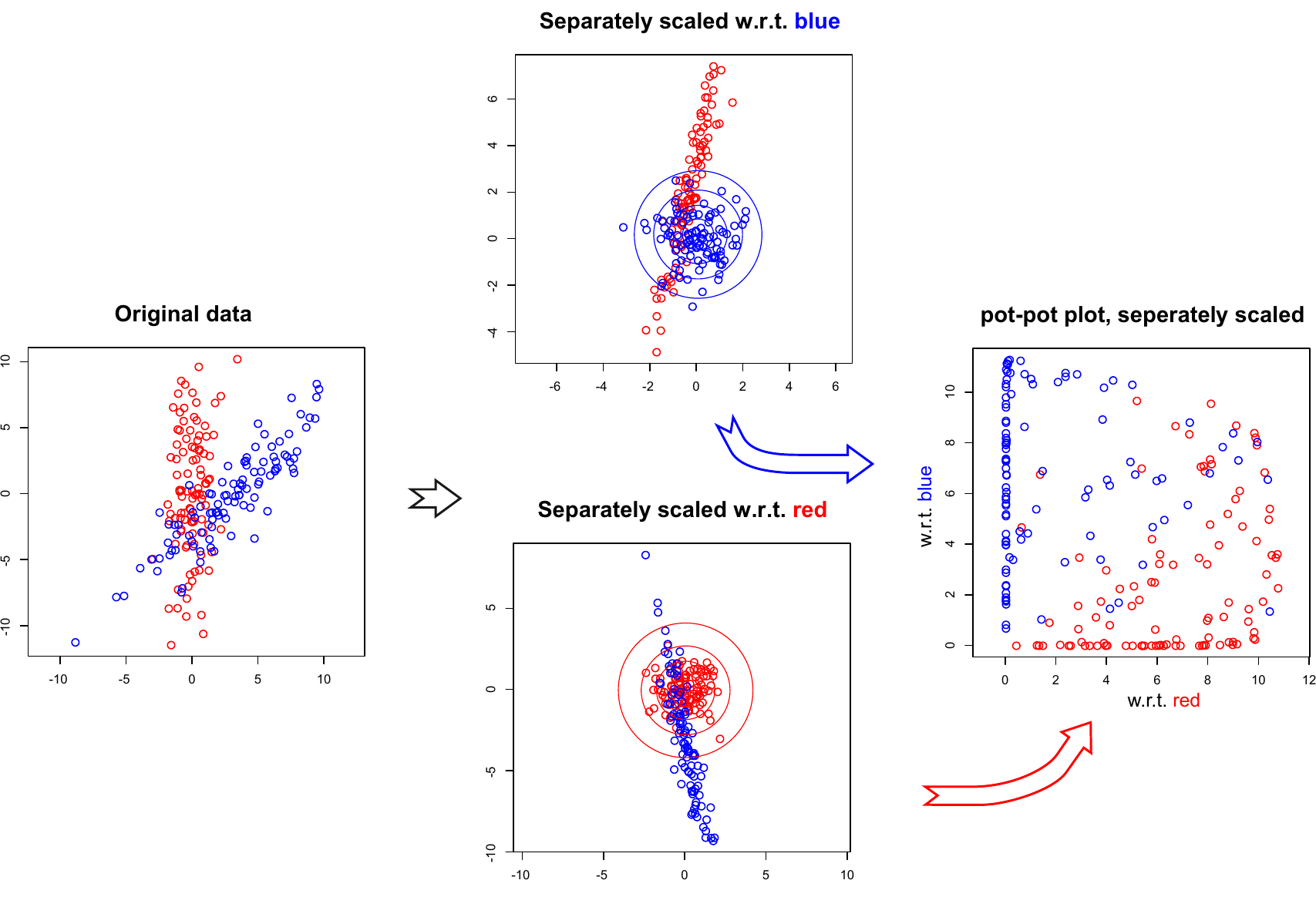}
    \caption{
        The data is separately scaled.
        The plots show the original data, the scaled data with their lines of equal potential,
                        and the corresponding \mbox{pot-pot plot}.
    }
    \label{f:scaling_separately}
\end{figure}

An alternative is \textit{separate scaling}. It results in potentials
\[\phi_j(\bmx)= p_j \hat f_j(\bmx) = p_j \frac 1{n_j} \sum_{i=1}^{n_j}{K_{h_j^2 \hat \bmSigma_j}(\bmx-\bmx_{ji})}\,.
\]
With separate scaling the two kernels are built with different bandwidth matrices $\bmH_j = \bmC_3 = h_j^2\hat \bmSigma_j$ to fit the form of each class. We need estimates of two covariance matrices and have two parameters, $h_1^2$ and $h_2^2$, to tune.
Figure \ref{f:scaling_separately} illustrates the approach. It is clearly seen that many points receive much less potential with respect to the opposite class. 

In case of heavy-tailed data, robustness is achieved by applying the Minimum Covariance Determinant (MCD) or the Minimum Volume Ellipsoid (MVE) estimates to transform the data.
Note, that in some cases the  problem may occur, that the covariance matrix is singular, e.g. if the dimension is higher than the number of points. In this case one may use a proper pseudoinverse.

As tuning the parameters comes at high computational costs, we try to simplify the tuning in the case of separate scaling.
Either we use the same parameter, $h_1^2=h_2^2\,$ for both classes or we establish
some relationship between the two parameters, $h_2^2 = g(h_1^2)\,$ where $g$ is a function of the first bandwidth.
After a proper function $g$ is found only one parameter must be tuned.

In our experiments (sec. 7), we observe a relationship between the bandwidth parameters {that provide the smallest error rates}.
  For the real data sets we see that, with separate scaling, close to smallest classification errors are usually achieved on a straight line in the $(\log_{10}h^2_1,\log_{10}h^2_2)$-plane (see Fig. \ref{f:exp:errorplots_real} and the description there).
We profit from this observation and spare the effort of separately tuning the two parameters.
Note that the line is not always the main diagonal.
We propose to regress one parameter on the other, evaluating the error at a few pairs $(\log_{10}h^2_1,\log_{10}h^2_2)$ and using them as sampling points.
Specifically, we calculate the error at five sets of five points, with the five-point sets being taken orthogonally to the main diagonal of the plot; cf.\ Figures \ref{f:exp:errorplots_sim} and \ref{f:exp:errorplots_real}. Then the minimal error is found in each set and a linear or non-linear regression of $\log_{10}h_2^2$ on $\log_{10}h_1^2$ is used to find a proper relation between the bandwidths.
{Consequently, we combine separate scaling with a linear bandwidth regression function, $g$, which simplifies the procedure enormously.}
Specifically, we use $g$ to determine $h^2_2$ for every $h^2_1$, cross-validated at 60 possible values, while the full tuning would involve cross-validation of $(h^2_1, h^2_2)$ at 3600 points.
Clearly, separate scaling with bandwidth regression yields the same computational complexity as joint scaling.

\section{Experiments}\label{sec:experiments}

We have conducted an experimental survey to check the theoretical implications and to compare the performance of the proposed method with several traditional classifiers and $DD$-classification using popular global depths.

In the experiments we consider two classes. We compare joint and separate scaling of the classes and examine the variants of bandwidth selection.
As Figures \ref{f:exp:errorplots_sim} and \ref{f:exp:errorplots_real} illustrate, the error functions are multimodal and erratic, and can hardly be minimized in a way other than iterative search. In our experiments we select the bandwidth within a wide range and use logarithmic steps.

\subsection{The data}\label{ssec:exp:thedata}

Simulated as well as real data are considered in the experiments.
The first two simulated series consist of two-dimensional data sets of two normally distributed classes. The classes are located at different distances, and they are scaled and rotated in different ways:
\begin{enumerate}
\item Location: \tabto{1.8cm} $C_1 \sim N\bigl({0\brack0}{1~0\brack0~1}\bigr),$ $C_2 \sim N\bigl({l\brack0}{1~0\brack0~1}\bigr),$ $l=1,2,3,4;$\smallskip
\item Scale:    \tabto{1.8cm} $C_1 \sim N\bigl({0\brack0}{1~0\brack0~1}\bigr),$ $C_2 \sim N\bigl({3\brack0}{1~0\brack0~s}\bigr),$ $s=1,2,3,4,5;$\smallskip
\item Scale*:    \tabto{1.8cm} $C_1 \sim N\bigl({0\brack0}{1~0\brack0~1}\bigr),$ $C_2 \sim N\bigl({3\brack0}{s~0\brack0~1}\bigr),$ $s=2,3,4,5;$ \smallskip
\item Rotation:   \tabto{1.8cm} $C_1 \sim N\bigl({0\brack0}{1~0\brack0~5}\bigr),$ $C_2 \sim N\bigl({3\brack0}{1~0\brack0~5}\bigr).$\\
        Firstly, $C_1$ and $C_2$ are generated.
        After that $C_2$ is rotated around \\ $\mu_2=(3,0)$ by $\alpha \in [0, \pi/2]$ in 5 steps, then $C_2$ stays rotated by $\alpha = \pi/2$ and $C_1$ is rotated around $\mu_1=(0,0)$ by the same angles, giving 9 data sets in total.
\end{enumerate}

The training sequence of the first series contains 100 points in each class, while the second is more asymmetric and contains 1000 resp.\ 300 points in the two classes.
The testing sequence contains 300 points in each class of the first series, and 1000 resp.\ 300 points in two classes of the second series.
We use the following acronyms for the data sets: the number of series ($1$ for equally, $2$ for unequally sized classes), the name of transformation, the number of the transformation. E.g., \verb"2scale2" means a data set with 1000 and 300 points in the classes, transformed by `scale' transformation with $s=2$.

The third simulated series \citep{DuttaCG12} is generated with uniform distributions on nested disks.
{In this case the classes cannot be separated by a simple line or curve.} 
The two classes are distributed as $C_1 \sim U_d(0,1)+U_d(2,3)$ and $C_2 \sim U_d(1,2)+U_d(3,4),$
with $U_d(r_1,r_2)$ being the uniform distribution on $\{\bmx\in\mathbb R^d:r_1<||\bmx||<r_2\}$.
We generate the data sets in two ways: The first ones, as proposed by \cite{DuttaCG12}, have an equal number of points in both classes ($n_1 = n_2 = 100$ resp.\ $n_1 = n_2 = 400$). Note that in this case one class is less densely populated than the other. The second ones are generated so that the points of both classes are equally dense, with $n_1 = 80$, $n_2 = 120$ resp.\ $n_1 = 300$, $n_2 = 500$. The naming of these data sets reflects the number of points in the classes.

For the simulated data sets the classification errors are estimated using training and testing sequences drawn from the same distributions. The procedure is replicated 40 times and the mean values are taken. The standard deviations of the error rates are mostly around five to ten times smaller than their values, and this ratio is smaller than two only in one percent of all cases.

We also use 50 real multivariate binary classification problems, collected and described by \cite{MozharovskyiML15}.
The data sets are available at \url{http://www.wisostat.uni-koeln.de/de/forschung/software-und-daten/data-for-classification/} and in the R-package `ddalpha'. The data have up to 1\,000 points in up to 15 dimensions; they include asymmetries, fat tails and outliers.
As the real data sets have no separate testing sequence, the cross-validation procedure described in the introduction is used to estimate the classification errors. We exclude one or more points from the data set, train the classifiers on the remaining points and check them using the excluded points. The number of excluded points is chosen to limit the number of such iterations to 200.


\subsection{Comparison with depth approaches and traditional classifiers}\label{ssec:exp:compdepth}

We compare the classifiers with the Bayes classifier for the simulated data and with LDA for the real data. Note that in this case LDA gave the best results among the traditional classifiers: \textit{linear discriminant analysis} (LDA), \textit{quadratic discriminate analysis} (QDA), and \textit{$k$-nearest neighbors} (\kNN{}).
We introduce an efficiency index as the ratio of the error rates of the chosen classifier and the referenced one (the Bayes classifier or the LDA, resp.):
$I_{classifier} = {\epsilon_{classifier}}/{\epsilon_{reference}}$.
The index measures the relative efficiency of a classifier compared to the referenced one for each particular data set.
{We study the distribution of the efficiency index for each method over all data sets using box plots, which allows to compare the efficiency of different methods visually.}

We compare our method with \emph{kernel density estimation} (KDE) and \emph{$DD$-plot classification}.
To construct the $DD$-plots, we apply five most popular global depth functions, that can be calculated in reasonable time: zonoid, halfspace, Mahalanobis, projection and spatial (=$L_1$) depth. We do not use simplicial or simplicial volume depth, as they need much more calculation time, which is not feasible in higher dimensions. For details on these depth notions, see \cite{ZuoS00}, \cite{Serfling06}, and \cite{Mosler13}.
Then we use $\alpha$-classification \citep{LangeMM14a}, polynomial \citep{LiCAL12} or \kNN{}-classification on the $DD$-plot, or take its diagonal as the separation line.
The diagonal separation on the $DD$-plot corresponds to the maximum-depth approach.
In contrast to this, the $\alpha$-classifier as well as the polynomial separator and the \kNN{}-method produce a non-linear separation of the $DD$-plot.

The zonoid and halfspace depths vanish outside the convex support of the training data and the points lying there have zero depth. If a point has zero depth w.r.t.\ both classes it is called an \textit{outsider}; it cannot be classified on the $DD$-plot. \cite{LangeMM14a} propose a separate outsiders treatment procedure, which classifies outsiders in the original space.
However, if an outsider treatment procedure is used, the obtained error rate is a mixture of that of the $DD$-classifier
and that of the outsider treatment procedure.
{In our study we use QDA as an outsider treatment procedure for the simulated data and LDA for the real data.}
Note that the number of outsiders is much larger in the real data sets than in the simulated data. It is higher than 50\% for $2/3$ of the real data sets.

{We also add the depth-based \kNN{} of \cite{PaindaveineVB15} for comparison, as it is close in spirit to depth-based classification and to \kNN. Here we apply halfspace and Mahalanobis depths, and choose $k$ by cross-validation.}

On the pot-pot plot we proceed in a similar way. We separate the classes by either the diagonal line or the $\alpha$-classifier or \kNN.
Clearly, using the diagonal as a separator corresponds to KDE applied to the original data. This is combined with different bandwidth approaches:
(1) joint scaling and optimizing a single $h^2$, (2) separate scaling and optimizing both $h_1^2$ and $h_2^2$, (3) separate scaling, regressing $h_2^2$ on $h_1^2$ (where $h_1^2$ belongs to the larger class), {and optimizing the regressor only}; see Section \ref{ssec:exp:bandwidth}.

Tables \ref{t:errors_sim_depths} and \ref{t:errors_depths} show errors from using the different methods.
The methods are grouped by type, and the best values within each group are marked black. The best classifiers for the particular data set are marked red.
For the $DD$- and pot-pot classifiers we report the errors of the $\alpha$-classifier, averaged over the replications resp.\ cross-validation runs. For the pot-pot classifiers we use the errors, obtained with the best bandwidth parameters, and report them for the joint, separate and regressive separate 
approaches. The error rate estimating procedure is stable as the standard errors are very small (not reported in the tables).
Figure \ref{f:exp:boxplots} exhibits box plots of the efficiency index of each of these methods.
Here we use the $\alpha$-classifier to display the efficiency of different methods.
The tables and figures for other separating methods (diagonal, polynomial and \kNN{}) are transferred to a separate Online Appendix.

The experimental results show that the proposed method outperforms all compared methods on the real data and shows competitive results on the simulated data. We also observe that separate scaling (\emph{pot-pot separate}) is more efficient than joint scaling (\emph{pot-pot joint}). Under the name \emph{pot-pot regressive separate} we show results that are obtained with separate scaling and bandwidth regression.

Tables \ref{t:errors_sim_potential} and \ref{t:errors_potential} compare the minimal errors obtained with pot-pot classifiers that use the three separating procedures on the pot-pot plot: diagonal, $\alpha$, and \kNN. Also additional
bandwidth approaches in estimating the potentials (\textit{ROT}, {\textit{mM}}) are included; see Section \ref{ssec:exp:bandwidth} below.
Figure \ref{f:exp:boxplots_potential} illustrates the corresponding boxplots of the efficiency index.
For the real data sets, using either $\alpha$- or \kNN{}-classifiers on the \mbox{pot-pot} plot shows better classification results than classical KDE does.  The $\alpha$-classifier usually has a lower error rate than \kNN, but sometimes \kNN{} produces unexpectedly good results, as for example in the `cloud' and some of the `crab' data sets.
The error of the maximum-depth classifier slightly outperforms that of the $\alpha$- and \kNN{}-classifiers for the simulated data, but is more dispersed in the case of separate scaling.

The efficiency of the $DD$-plot classifiers is also compared for each of the five depths.
$DD\alpha$-, polynomial and \kNN{}-classifiers are more efficient than the maximum-depth classifier, and $DD\alpha$ shows best results in most of the cases (see Fig. \ref{f:exp:boxplots_DD_global}). We also observe that $DD\alpha$ shows almost the same results as the polynomial classifier, and slightly outperforms it for the real data.

The pot-pot approach performs much better than the other classifiers on real data. For the first two sets of simulated data, as they are generated by normal distributions, QDA and KDE are best classifiers under separate scaling (see \emph{pot-pot separate diagonal}).

\begin{figure}[h]
  \begin{center}
    \includegraphics[keepaspectratio=true,width = \textwidth, trim = 3mm 2mm 0mm 2mm, clip, page = 1]{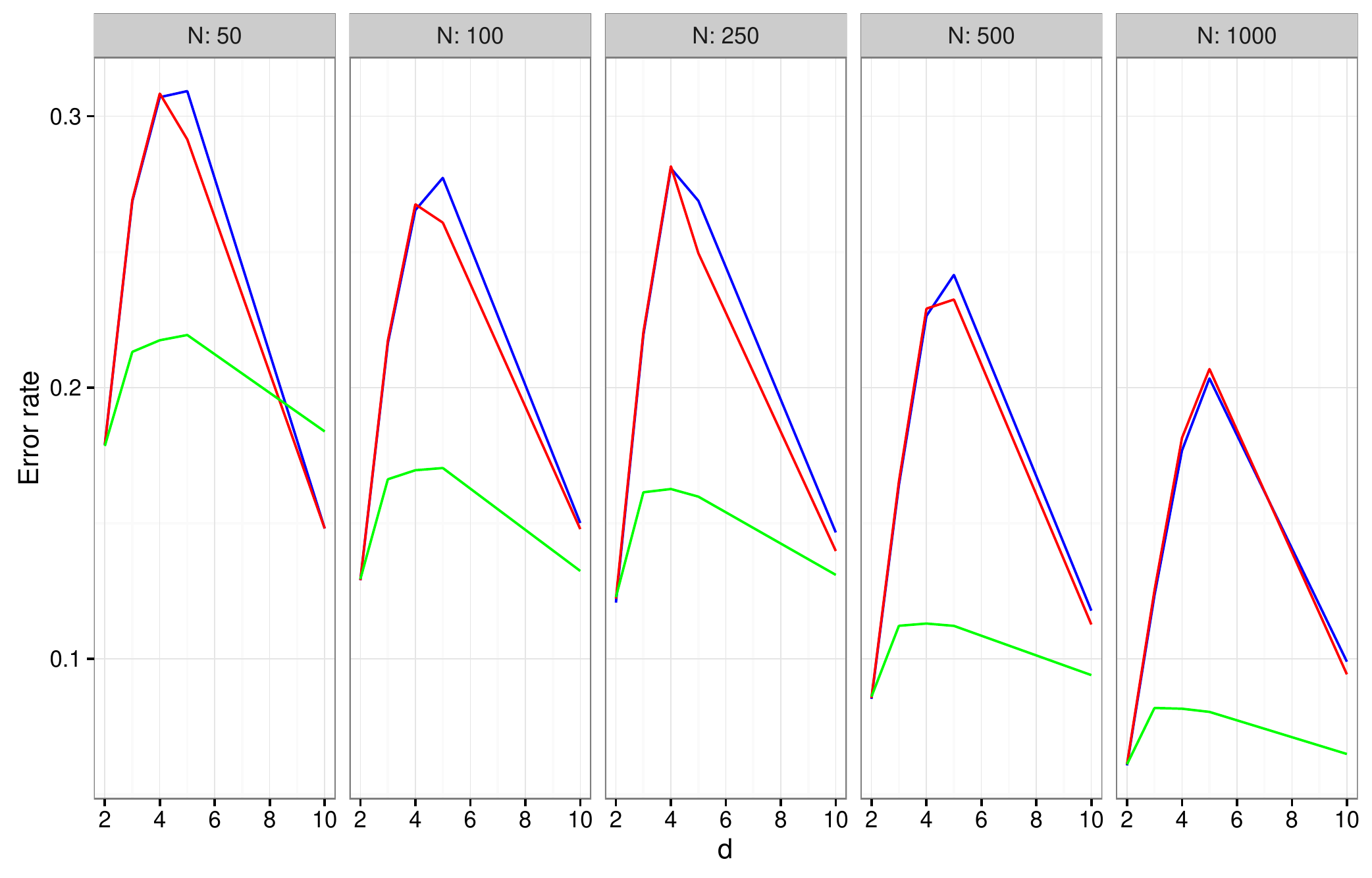}\\
     {\small {\color{blue}\LBL} diagonal,  {\color{green}\LBL} \kNN{}, {\color{red}\LBL} $DD\alpha$}
    \caption{Performance of the KDE and the pot-pot classifiers in the multidimensional space.}
    \label{fig:multidimens}
  \end{center}
\end{figure}

To illustrate the performance of the pot-pot classifiers in the higher dimensions we simulated multidimensional hyperspheres similar to the third simulated series in Section \ref{ssec:exp:thedata}. The points were generated uniformly in $d\in\{2, 3, 4, 5, 10\}$ with and sample length $n\in\{50,100,250,500,1000\}$.
With the growth of dimension the volume of the outer spheres increases faster than of the inner ones. In this case the classes become unbalanced and the probability of the first class is \{0.38, 0.31, 0.26, 0.21, 0.06\}, respectively. To make the classes equal we inverted the labeling in one half of the hypersphere.
We observe that with the growth of $d$ the error rate of the diagonal separation (=KDE) and the $DD\alpha$ grow much faster than the error rate of \kNN{}, see Figure~\ref{fig:multidimens}.
As shown before, in dimension 10 the hypersphere is divided into two halves, each containing mostly one class, and therefore the error rates of all classifiers become smaller. Nevertheless it is still much lower for \kNN{}, which means that the pot-pot plot allows to improve the separation even in higher dimensions.
As for the real examples, the advantage of the pot-pot classifiers is bigger in small dimensions, while in the dimensions higher than eight they mostly perform as good as the diagonal separation, see Table~\ref{t:errors_potential}.

\subsection{Selection of the optimal bandwidth}\label{ssec:exp:bandwidth}

We vary the kernel bandwidth parameters over a wide range using logarithmic steps.
The bandwidth selection process is shown on the bandwidths-to-errors plots, which illustrate the dependencies between the selected kernel bandwidths and the classification errors for particular data sets using diverse $DD$-classifiers under joint and separate scaling.
See Figures \ref{f:exp:errorplots_sim} and \ref{f:exp:errorplots_real} and the explanations there.
In the {\emph{joint scaling}} case the abscissa represents the logarithm of the bandwidth parameter $\log_{10}h^2\,$, and the ordinate the error rate. For the {\emph{separate scaling}} case the axes present the $\log_{10}h_i^2$ bandwidth of the first and the second classes kernels, respectively. The colors correspond to the classification errors achieved with these bandwidth combinations, where {\it red} indicates the highest error rate, {\it violet} the lowest, and the colors in between are in the rainbow order.

Experimentally we have found higher and lower bounds for the kernel bandwidth parameters search. The lower bound of $h^2 = 10^{-3}$ is explained with computational limitations, as the potential induced by such narrow kernels is too small to be represented with the machine type `double'. Thus, most points cannot be classified on the pot-pot plot, as they obtain zero potential. Such points are \emph{outsiders} in $DD$-classification, where they are classified separately; see \cite{LangeMM14a}. When the  bandwidth reaches the level of $h^2 = 10^3$, separation does not improve any more, since the kernels become too flat. As the classification error stabilizes at this level, we take it as an upper bound.

 It is also observed that extremely wide or narrow kernels give fairly good separation errors. This feature may be used for a fast `draft' estimation of the classification error which could be reached with the classifier, and possibly for reducing the search intervals of the bandwidth parameters.

In selecting the kernels' bandwidths we compare the performance of the generalized Scott's rule of thumb (column \emph{ROT}) and the extreme bandwidths (column \emph{mM}). The latter means that we use the bound (either lower or upper) of $h^2$ that gives the smaller error.
The rule of thumb works better for the simulated data sets than for the real ones.

For any pair of normally distributed classes (shifted, scaled, and/or rotated)
the optimal bandwidths are equal for both classes $h_1^2 = h_2^2$; they lie around $h^2 = 1$ if the classes have about the same size. This also holds for the $DD\alpha$- and \kNN{}-classifiers if the classes have different sizes, while for the maximum-depth classifier a shifted line is observed.
The results obtained using the rule of thumb are close to the optimal ones.

\begin{figure}[h]
    \centering
    \includegraphics[keepaspectratio=true,width = 0.49\textwidth, page = 1]{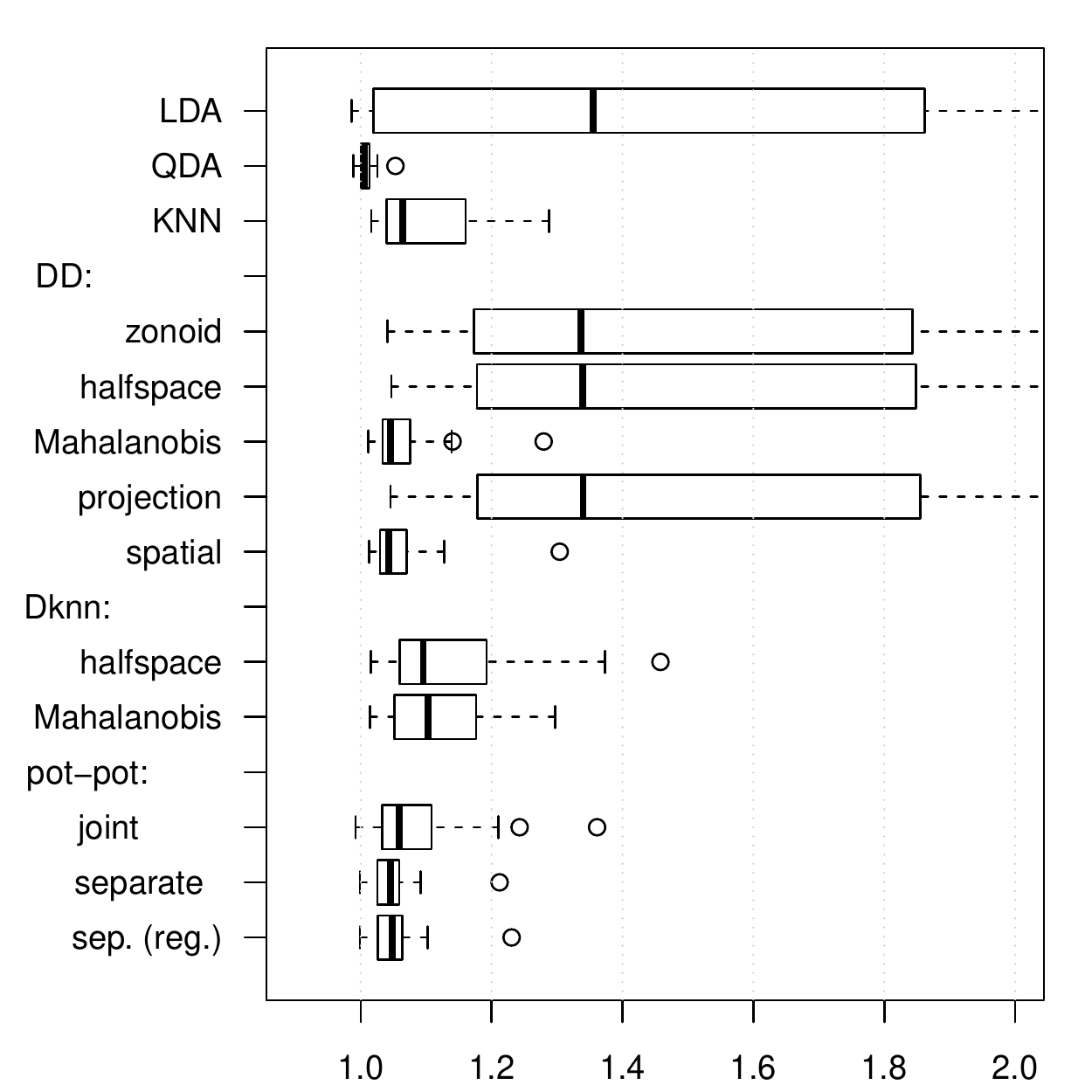}
    \includegraphics[keepaspectratio=true,width = 0.49\textwidth, page = 1]{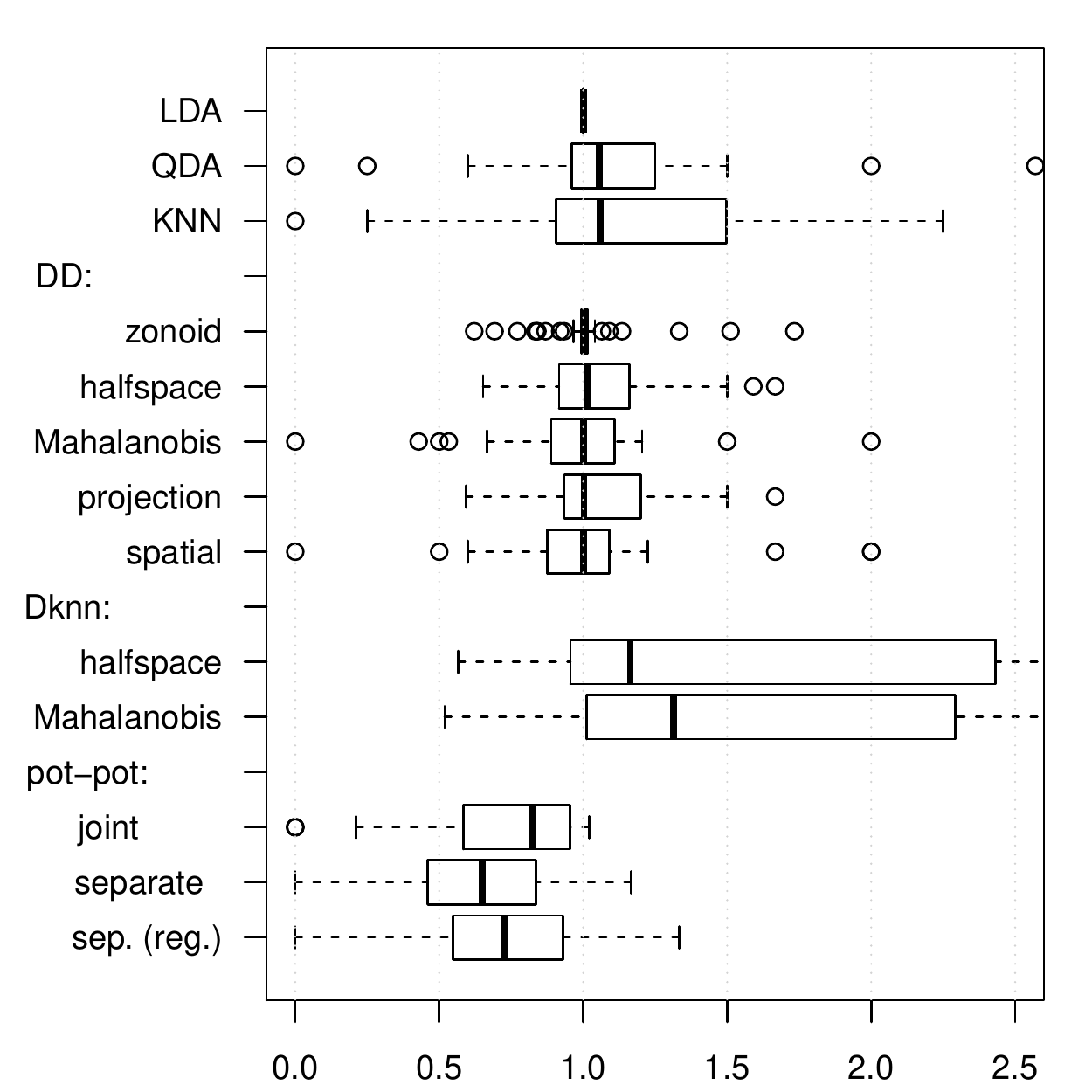}

        a) simulated elliptical data ~~~~~~~~~~~~~~~~~~~~~~~~~~~~~    b) real data
    \caption{
        Efficiency of the methods. For the $DD$- and pot-pot classifiers the errors of the $\alpha$-classifier are given.
    }
    \label{f:exp:boxplots}
 \parbox[t]{\textwidth}{ \footnotesize
 \medskip
    \tabto{0.5 cm}
    The index is the relation of the error rates of the chosen classifier and the reference classifier.
    Here and in the following figures we take the Bayes risk as the reference for the simulated data and LDA -- for the real data.
    The index measures the relative efficiency of a classifier compared to the reference for a particular data set (the more efficient classifier has smaller index). For each classifier a boxplot is built that illustrates the distribution of the efficiency index over all data sets.
 }

        \bigskip
        \bigskip

    \centering
    \includegraphics[keepaspectratio=true,width = 0.49\textwidth, page = 1]{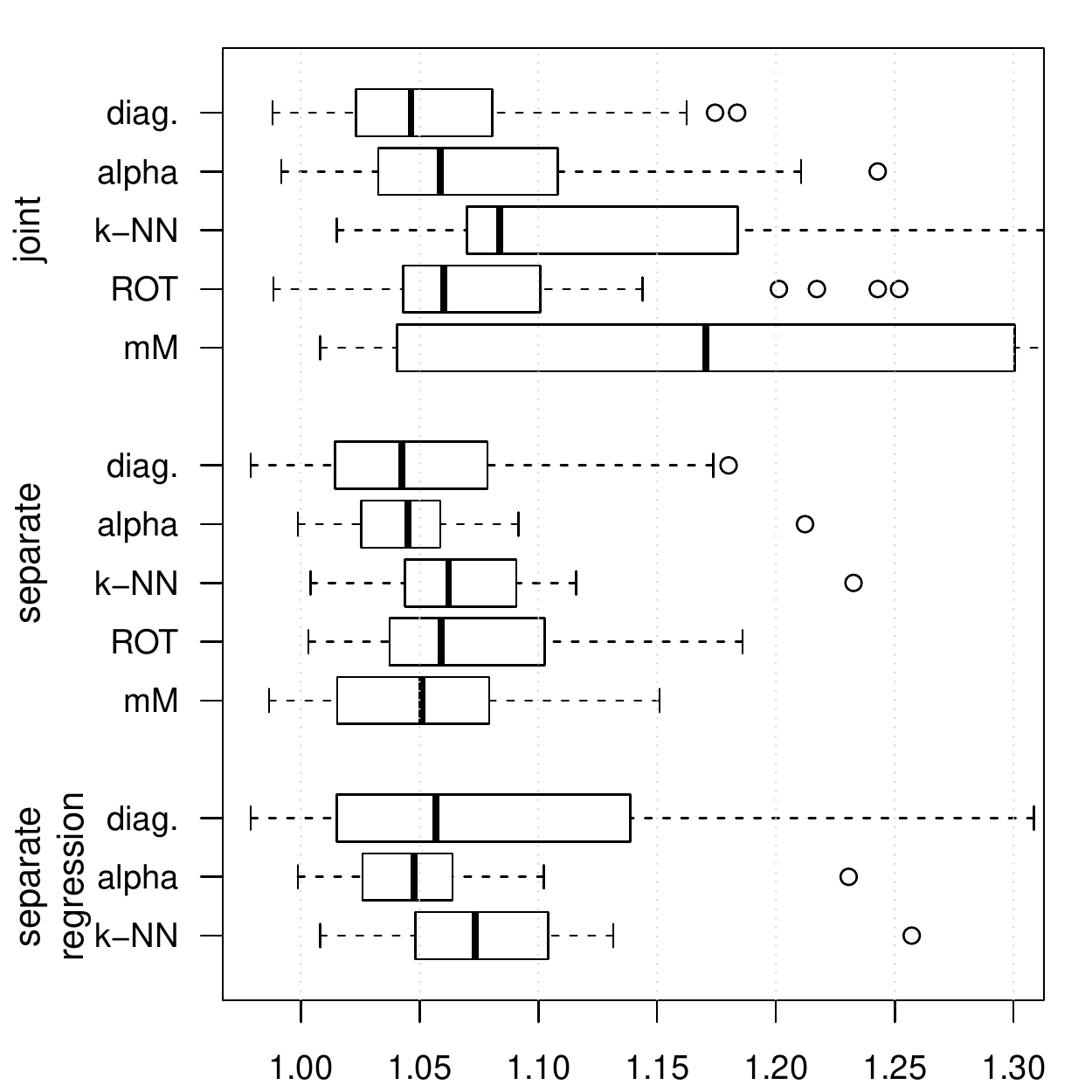}
    \includegraphics[keepaspectratio=true,width = 0.49\textwidth, page = 1]{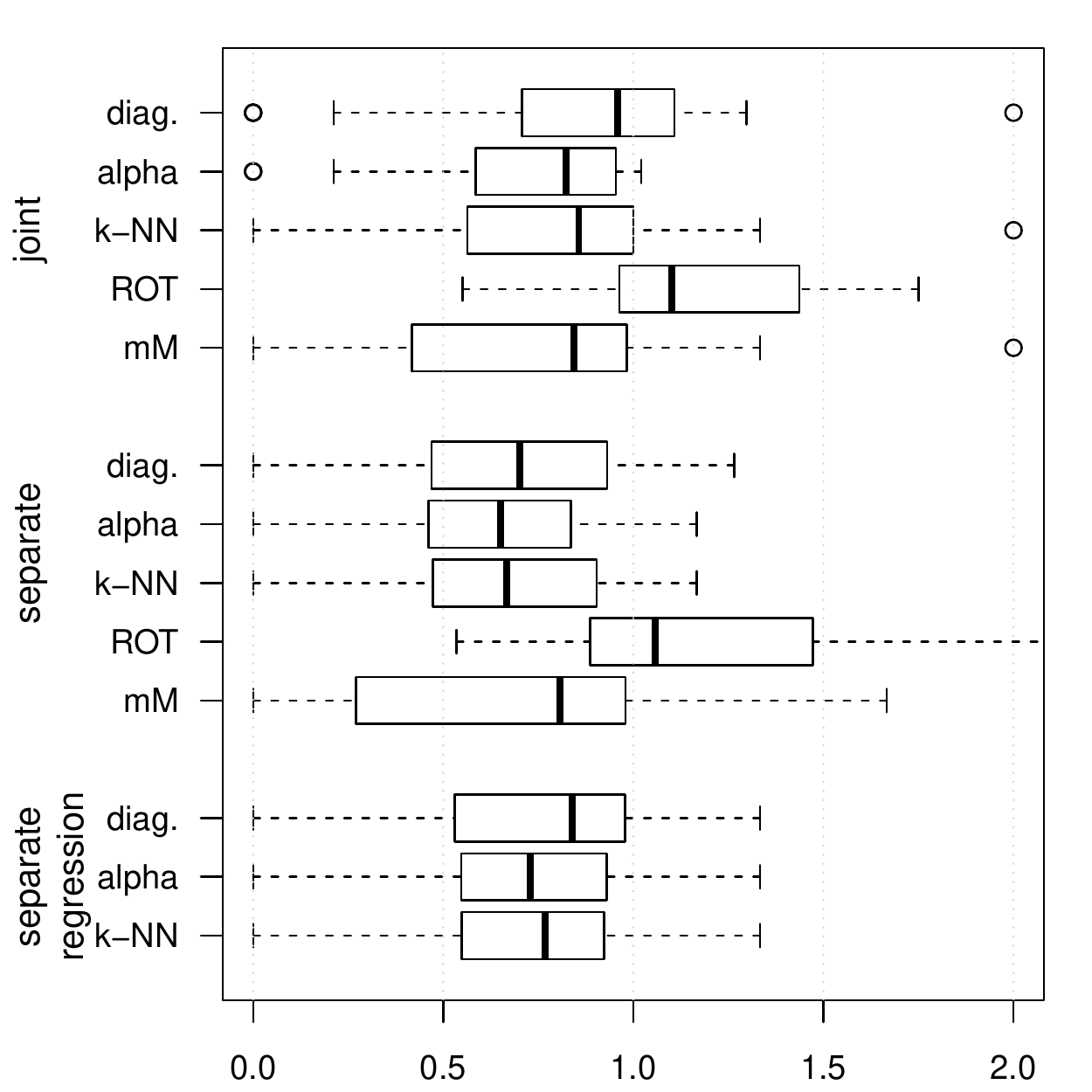}

        a) simulated elliptical data ~~~~~~~~~~~~~~~~~~~~~~~~~~~~~    b) real data
    \caption{
        Efficiency of the pot-pot classifiers.
    }
    \label{f:exp:boxplots_potential}
\end{figure}

We compared the results of the full-bandwidth tuning and the regression approach, described in the end of section \ref{sec:scaling}.
The linear regression is the most reasonable, as higher order regressions, described under Figure \ref{f:exp:errorplots_sim} did not improve the efficiency of our procedure relative to simple linear regression, which is illustrated in Figure \ref{f:exp:boxplots_regressions}.
This bandwidth regression approach is abbreviated as \emph{pot-pot regressive separate} in the tables and figures.
Observe that the results are close to the minimum obtained by full bandwidth search using separate scaling.
 They also outperform the ones provided by joint scaling, as Figures \ref{f:exp:boxplots}, \ref{f:exp:boxplots_potential} and \ref{f:exp:boxplots_regressions} demonstrate.
This approach works much better on the real data settings, while on simulated data the difference between joint and separate scaling is not really large.  Note that in this approach the $\alpha$-classifier shows best results for both simulated and real data.

\subsection{Comparison of the classification speed}
As the experiments have shown, the proposed method has a very good relative performance, but at the cost of a reduced training speed.
The \emph{training time} of a pot-pot (or $DD$-) classifier consists of the time to calculate the potential (resp. the depth) of each point w.r.t.~each class and of the time needed to train the separator; the latter does virtually not depend on the choice of the space transformation function. If $q>2$ and the aggregating procedures are involved, multiple separators may be trained on the pot-pot plot.
The \emph{classification time} contains only the time for calculation of the potentials (resp. depths) of a given point w.r.t.~each class.

\begin{figure}[h]
  \begin{center}
    \includegraphics[keepaspectratio=true,width = \textwidth, trim = 3mm 2mm 0mm 2mm, clip, page = 1]{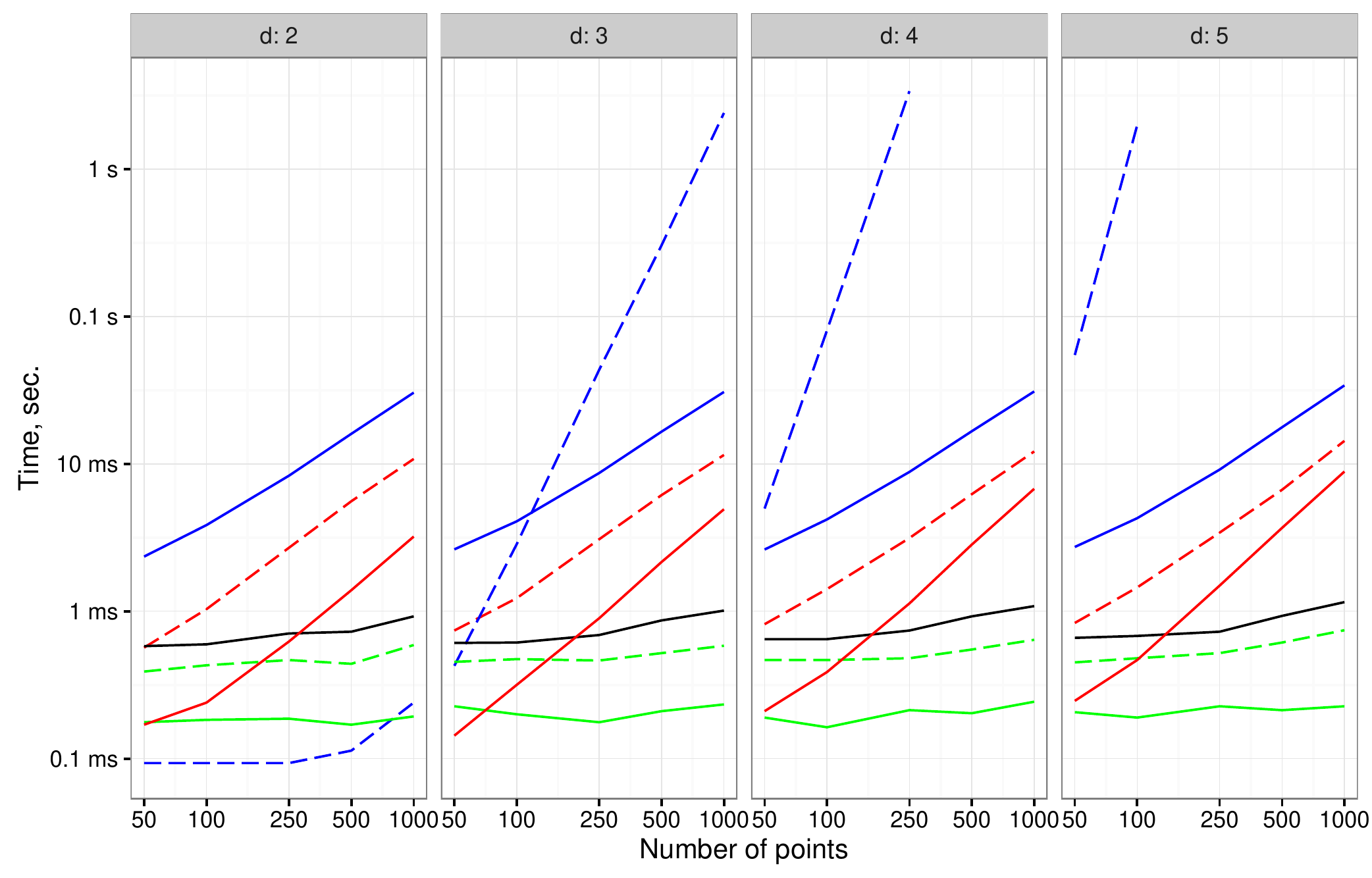}\\
     {\small {\color{black}\LBL} potential, {\color{red}\LBL} zonoid, {\color{red}\LBD} halfspace,

     {\color{green}\LBL} Mahalanobis, {\color{green}\LBD} spatial, {\color{blue}\LBL} projection, {\color{blue}\LBD} simplicial}
    \caption{Calculation time of a single data point for various depth functions and potential, on the logarithmic time scale.}
    \label{fig:depth_speed}
  \end{center}
\end{figure}

We compare the computation times of potentials and various depth notions by graphics in Figure~\ref{fig:depth_speed}.
On the logarithmic time scale, the lines represent the time (in seconds) needed to compute depth or potential of a single point, averaged over 50 points w.r.t. 60 samples, varying dimension $d\in\{2, 3, 4, 5\}$ and sample length $n\in\{50,100,250,500,1000\}$.
Due to the fact that computation times of the algorithms do not depend on the particular shape of the data, the data has been drawn from the standard normal distribution. Some of the graphics are incomplete due to excessive time.
Projection and halfspace depths have been approximated using 1\,000 random projections and simplicial depth (if $d>2$) has been approximated using 5\% of simplices. The other depths have been computed exactly as well as simplicial depth for $d=2$. The calculation of simplicial depth dramatically slows down with the growth of $n$ and $d$, therefore we do not include it in the comparison of Section \ref{ssec:exp:compdepth}.

We observe that it takes from 0.5ms to 1ms to calculate the potential of one point. The calculation speed does not depend on the dimension of the data and slightly depends on the number of points. In big data sets it is only outperformed by Mahalanobis and spatial depths.

This advantage is nevertheless suppressed by the tuning of the bandwidth parameters.
Having $k_p$ bandwidth parameters the error rate obtained with each of them is estimated by cross-validation that repeats the training and classification procedures for $k_e$ times.
A usual choice of $k_e$ is 10, but in this paper we set it to 200 to obtain precise estimates.
The number $k_p$ of possible values of the bandwidth parameter is set to $60$ in the {\emph{joint scaling}} case, $3600$ in the {\emph{separate scaling}} case, and $25+60$ in the \emph{regressive separate} approach.
This means that with the regressive separate approach we have trained the pot-pot classifier  $85*200=17\,000$ times, and the $DD$-classifiers only 200 times to estimate the error rate. In practice these numbers may be seriously reduced by iterating less bandwidth parameters and shortening their range, and applying less iterations during cross-validation. Note that the classical KDE classification needs the same number of iterations as the pot-pot classifiers to tune the bandwidth parameters, given that the same bandwidth matrices are taken in both approaches.

\section{Conclusion}\label{sec:conclusion}

A new method is proposed that combines ideas of kernel discriminant analysis and depth-depth-plot classification.
Potentials of data points, which amount to weighted kernel-estimated densities, are used for classification on a potential-potential (pot-pot) plot.
Compared with classical approaches the method shows a very good relative performance, especially on real data settings.

The two most important aspects of the method of potentials are:
Firstly, compared to classification methods based on depths the method reflects local properties of the distributions and, thus, gives better results for multimodal and non-elliptical distributions.
Secondly, the use of the {pot-pot plot} allows for more sophisticated separations than the simple comparison of estimated densities.

Consequently, the bandwidth parametrization and the selection of bandwidth parameter values can be kept simple.
Instead of scaling the kernels we scale the data separately and apply a simple spherical kernel to these sphered data. Then the kernel bandwidth itself is tuned using just one parameter.
Joint and separate scaling of the classes are compared.
Under separate scaling the kernels provide a better fit to the classes' distributions, which clearly improves the classification.

\begin{figure}[h]
\section*{Appendix}
\begin{tabular}{p{0.3cm}c}
\hspace{-0.4cm}
\begin{adjustbox}{valign=t}
\rotatebox[origin=c]{90}{
\begin{tabularx}{0.68\textwidth}{YYYc}
\footnotesize
disks 100x100 & 2scale4 & 2rotate5 &
data
  \end{tabularx}
}
\end{adjustbox}

  &

  \hspace{-0.4cm}
  \begin{adjustbox}{valign=t}
  \begin{tabularx}{0.96\textwidth}{YYYY}
    \multirow{2}{*}{{Joint scaling}} & \multicolumn{3}{c}{Separate scaling}\\
  & {$DD\alpha$} & {\kNN{}} & {diagonal} \\
  \multicolumn{4}{l}{\hspace{-0.6cm}
                       \includegraphics[keepaspectratio=true,width = 1.02\textwidth,trim=0cm 0.5cm 0cm 0.5cm, clip=true, page = 1]{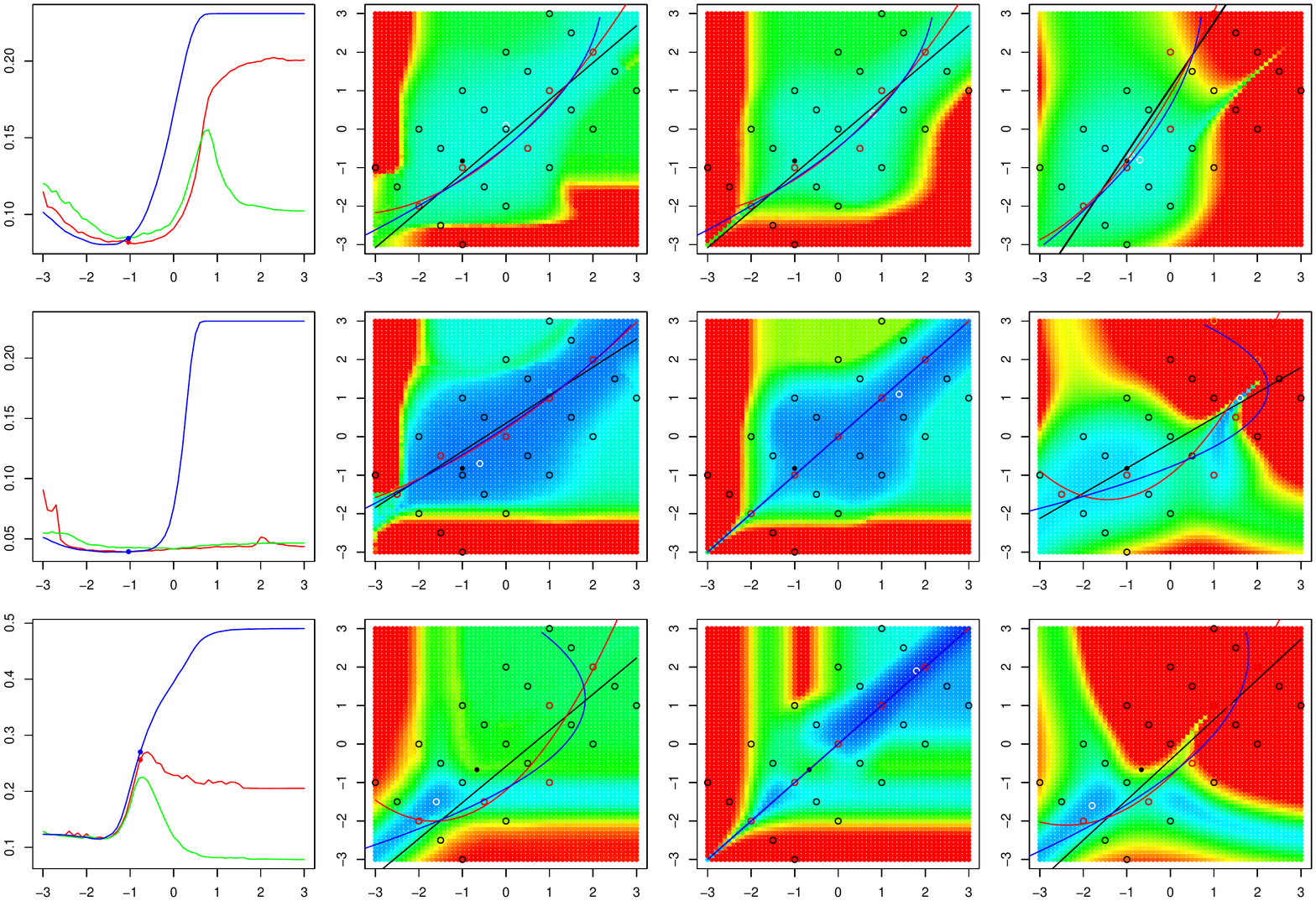}}
  \end{tabularx}
  \end{adjustbox}
   \end{tabular}
    \caption{
        Examples of bandwidths-to-error plots
        (simulated data).
    }
    \label{f:exp:errorplots_sim}
  \parbox[t]{\textwidth}{ \footnotesize
  \medskip
Left column of panels (joint scaling):
    \tabto{0.5 cm} abscissa $\widehat=$ $\log_{10}h^2$; ordinate $\widehat=$ classification error rate; \\
    \tabto{0.5 cm} classifiers:  diagonal (\blue{\it blue}), \kNN{} (\green{\it green}),  $DD\alpha$ (\red{\it red}).

    \smallskip
Other panels (separate scaling):
    \tabto{0.5 cm} abscissa $\widehat=$ $\log_{10}h_1^2$; ordinate $\widehat=$ $\log_{10}h_1^2$;
    \tabto{0.5 cm} colors: classification error rates from {\it violet} to {\it red};\\
    \tabto{0.5 cm} points: {\it black} -- sample points (grouped orthogonally to the main diagonal); \\
                \tabto{1 cm}  minima in each test group ({\it red points}); global minima ({\it white points});\\
    \tabto{0.5 cm} regressions ($h_1^2$ is the bandwidth of the larger class):\smallskip
            \newline \tabto{1 cm} \emph{{Linear}}            \tabto{4.3cm} --- $\log_{10}h_2^2 = a + b \log_{10}h_1^2$\,;\smallskip
            \newline \tabto{1 cm} \emph{{Quadratic}}         \tabto{4.3cm} \red{---} $\log_{10}h_2^2 = a + b \log_{10}h_1^2 + c  (\log_{10}h_1^2)^2$\,;\smallskip
            \newline \tabto{1 cm} \emph{{Quadratic inverted}}\tabto{4.3cm} \blue{---} $\log_{10}h_1^2 = a + b  \log_{10}h_2^2 + c  (\log_{10}h_2^2)^2$\,.

  \medskip
  See detailed description under Figure \ref{f:exp:errorplots_real}.
  }

\end{figure}

\begin{figure}[h]
\begin{tabular}{p{0.3cm}c}
\hspace{-0.4cm}
\begin{adjustbox}{valign=t}
\rotatebox[origin=c]{90}{
\begin{tabularx}{0.68\textwidth}{YYYc}
\footnotesize
segmentation & iris & biomed &
data
  \end{tabularx}
}
\end{adjustbox}

  &

  \hspace{-0.4cm}
  \begin{adjustbox}{valign=t}
  \begin{tabularx}{0.96\textwidth}{YYYY}
    \multirow{2}{*}{{Joint scaling}} & \multicolumn{3}{c}{Separate scaling}\\
  & {$DD\alpha$} & {\kNN{}} & {diagonal} \\
  \multicolumn{4}{l}{\hspace{-0.6cm}
                       \includegraphics[keepaspectratio=true,width = 1.02\textwidth,trim=0cm 0.5cm 0cm 0.7cm, clip=true, page = 1]{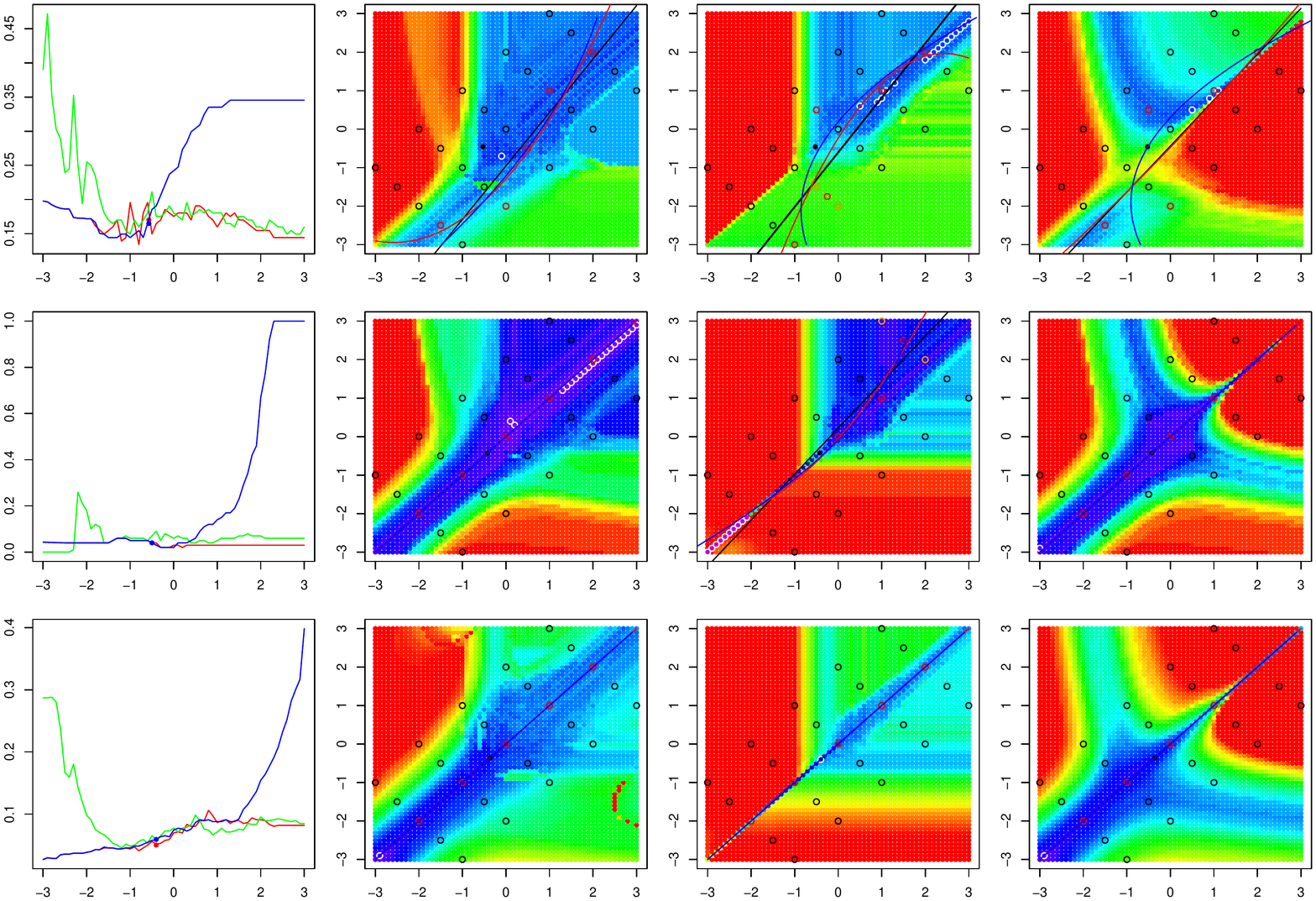}}
  \end{tabularx}
  \end{adjustbox}
\end{tabular}
    \caption{
        Examples of bandwidths-to-errors plots
        (real data).
    }

    \label{f:exp:errorplots_real}
\parbox[t]{\textwidth}{ \footnotesize
\medskip
    \tabto{0.5 cm}
    See the legend under Figure \ref{f:exp:errorplots_sim}.

    \tabto{0.5 cm}
    The bandwidths-to-errors plots illustrate the dependencies between the selected kernel bandwidths and the classification errors for particular data sets using diverse $DD$-classifiers using joint (left column of panels) and separate (other panels) scaling.

    \tabto{0.5 cm}
     In the \textbf{\emph{joint scaling}} case the abscissa represents the logarithm of the bandwidth parameter $\log_{10}h^2\,$, and the ordinate the error rate.
    The rule of thumb (\emph{ROT}) errors are shown in bold.

    \tabto{0.5 cm}
    For the \textbf{\emph{separate scaling}} case the axes present the $\log_{10}h_i^2$ bandwidth of the first and the second classes kernels, respectively. The colors correspond to the classification errors achieved with these bandwidth combinations, where {\it red} corresponds to the highest error rate, {\it violet} to the lowest, and the colors in between are in the rainbow order. The {\it black points} represent the rule of thumb (\emph{ROT}) bandwidths.

    \tabto{0.5 cm}
    We search for the relationship between the bandwidth parameters using regressions. At first we calculate the classification error rate at 25 bandwidth points divided into five sets orthogonal to the main diagonal. Then the minimum is found over each set and a regression is computed to find the proper relation between the bandwidths.
    We use the found relationship to estimate error rates along the regression line, iterating one bandwidth parameter and calculating the other.
    For comparing the performance of this approach (\emph{pot-pot regressive separate}), to joint and separate scaling; see Fig. \ref{f:exp:boxplots_potential}.

    \tabto{0.5 cm}
    The minimum errors are found in Table \ref{t:errors_sim_potential} for simulated and in Table \ref{t:errors_potential} for real data.
}
\end{figure}

\begin{figure}[h]
    \centering
    \includegraphics[keepaspectratio=true,width = 0.49\textwidth, page = 2]{s_boxplots_.pdf}
    \includegraphics[keepaspectratio=true,width = 0.49\textwidth, clip=true, page = 2]{boxplots_real_.pdf}

        a) simulated elliptical data ~~~~~~~~~~~~~~~~~~~~~~~~~~~~~    b) real data
    \caption{
        Efficiency of $DD$-classifiers.
    }
    \label{f:exp:boxplots_DD_global}

        \bigskip
        \bigskip

    \centering
    \includegraphics[keepaspectratio=true,width = 0.49\textwidth,trim=2cm 5cm 0cm 0cm, clip=true, page = 1]{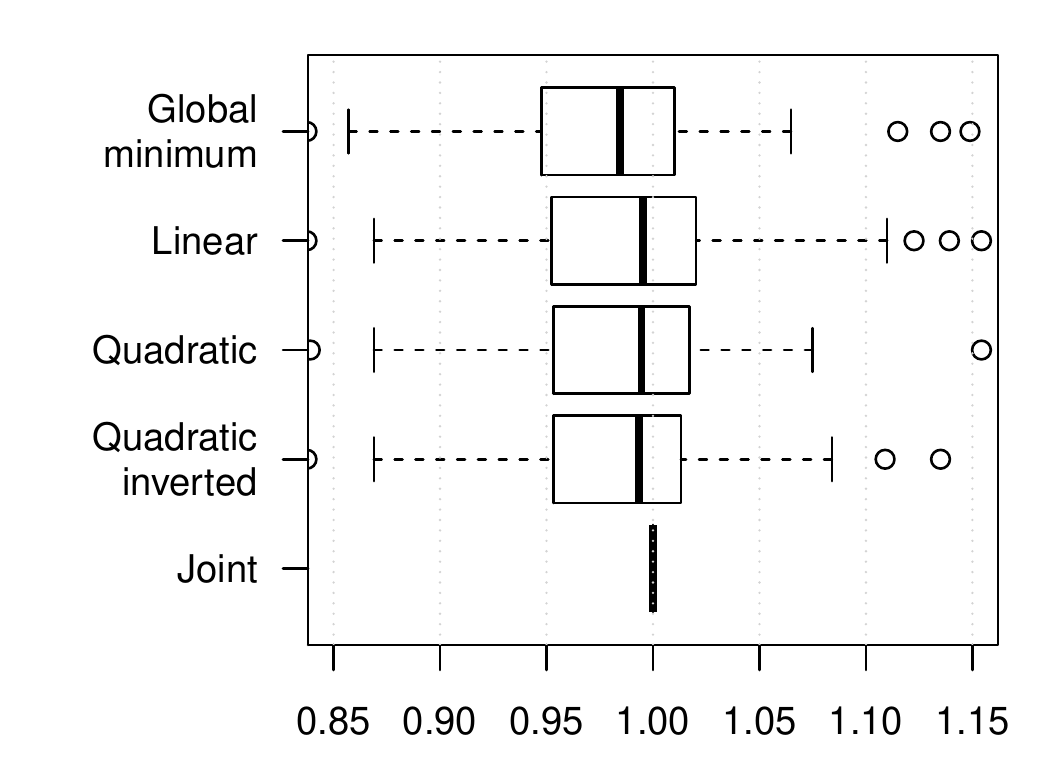}
    \includegraphics[keepaspectratio=true,width = 0.49\textwidth,trim=2cm 5cm 0cm 0cm, clip=true, page = 1]{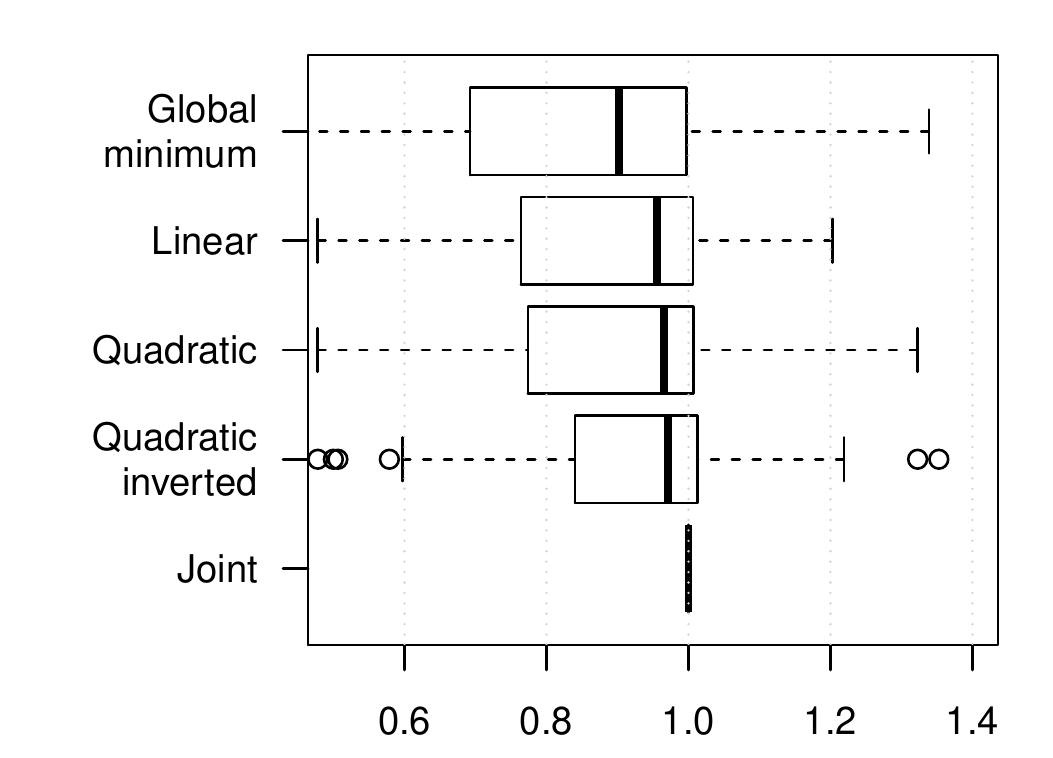}

        a) simulated elliptical data ~~~~~~~~~~~~~~~~~~~~~~~~~~~~~    b) real data
    \caption{
        Efficiency of separate scaling with different bandwidth regressions over joint scaling.
    }
    \parbox[t]{\textwidth}{ \footnotesize
    \medskip
        \tabto{0.5 cm}
        The boxplots show minimal errors using the same bandwidth regressions as in the description of Figure \ref{f:exp:errorplots_sim}.

        \tabto{0.5 cm} The first and the last rows show the global minima for resp. \emph{separate} and \emph{joint} scaling.

        \tabto{0.5 cm} The classifier using joint scaling is taken as the reference.
    }
    \label{f:exp:boxplots_regressions}
\end{figure}

As our experiments demonstrate, the classification error can be a multimodal and very erratic function of bandwidth parameters, which makes it difficult to minimize. We search for a global minimum by iterating over a proper set of possible parameter values (see Fig. \ref{f:exp:errorplots_sim} and \ref{f:exp:errorplots_real}).
Our experiments show  further that a roughly linear relation between the logarithms of two bandwidth parameters can be established under which a separation close to the best one is achieved. This allows us to restrict on tuning a single bandwidth parameter of one class $h_1^2$, using a linear regression to determine the bandwidth parameter of the second one $h_2^2$.
The bandwidth parameter is selected from the diapason $h^2 \in [10^{-3};10^{3}]$ with a logarithmic scale.

The naive KDE approach estimates the potential of a point regarding the classes and assigns the point to the class of maximum estimated potential.
Asymptotically the naive KDE approach reaches the optimal Bayes risk. However, with finite samples in higher dimensions these estimates are highly biased \citep{Friedman97} since the kernel bandwidth can only be roughly adjusted to the dependency structure of the distributions. In contrast our procedure is asymptotically Bayes-optimal as well, but reduces the influence of the finite-sample bias by additionally optimizing the line separating the potentials.

If two jointly scaled classes are considered, strong Bayes consistency is shown in general for pot-pot separation by \kNN{}, and under a slight restriction for separation by the $\alpha$-procedure.
Further consistency results may be derived for variants of this along the same lines. E.g.\ the data dependent kernels arising with separate scaling and $h_2$-$h_1$-regression are still continuous and regular; for consistency it is sufficient to warrant that the constant $\rho$ appearing in Theorem 10.1 of \cite{DevroyeGL96} satisfies $\rho=o(n)$.

The new method has been implemented as part of our R-package `ddalpha', which was also used for the experimental study.



\clearpage
\noindent
\parbox[t]{\textwidth}{
Tables \ref{t:errors_sim_depths} and \ref{t:errors_depths} show errors from using the different methods.
The methods are grouped by type, and the best values within each group are marked black. The best classifiers for the particular data set are marked red.

In the tables we use the following abbreviations for the data depths: HS for halfspace, Mah for Mahalanobis, Proj for projection and Spat for spatial. Dknn abbreviates the depth-based $k$-NN of Paindaveine and Van Bever (2015).
}
\begin{table}[h]
\centering
\caption{Error rates (in \%) of different classifiers for simulated data sets.}
\parbox[t]{\textwidth}{Columns (6) -- (10) and (13) -- (15): $\alpha$-classifier in the $DD$- and pot-pot plots.}
\smallskip
\label{t:errors_sim_depths}
\newcommand{\B}[1]{{\textbf{#1}}}
\newcommand{\R}[1]{{\red{#1}}}
\newcommand{\BR}[1]{{\red{\textbf{#1}}}}
\resizebox{1\textwidth}{!}
{
\begin{tabular}{|Hrr|rrr|rrrrr|rr|rrr|}
  \hline
  {H} & {(1)} & {(2)} & {(3)} & {(4)} &
 {(5)} & {(6)} & {(7)} & {(8)} & {(9)}         & (10)  & (11)
 & (12)  & (13)  & (14) & (15) \\
    {H} & {} & {} & {} & {} & {} &
 {} & {} & {} & {} & {}         & Dknn  & Dknn
 & pot-pot  & pot-pot  & pot-pot \\
  {H} & {dataset} & {Bayes} & {LDA} & {QDA} & {KNN} &
 {Zonoid} & {HS} & {Mah} & {Proj} & {Spat}         & HS & Mah
 & joint & separate & regress.\\
    {H} & {} & {} & {} & {} &
 {} & {} & {} & {} & {}    &     &   &
 &  &  & separate \\
  \hline
1 & 1dist1 & 30.9 & \BR{31.1} & 31.3 & 31.8 & 35.8 & 35.8 & \B{31.9} & 35.7 & \B{31.9} & 32.4 & 32.4 & 32.2 & \B{31.8} & \B{31.8} \\
  2 & 1dist2 & 15.8 & \BR{15.8} & 16.0 & 16.6 & 21.1 & 20.9 & \B{16.6} & 20.9 & 16.7 & 16.8 & 16.8 & 16.7 & \B{16.3} & 16.6 \\
  3 & 1dist3 & 6.7 & 6.9 & \BR{6.8} & 7.4 & 12.5 & 12.6 & \B{7.1} & 12.6 & 7.2 & 7.3 & 7.4 & 7.1 & \B{6.9} & \B{6.9} \\
  4 & 1dist4 & 2.0 & \BR{2.1} & 2.2 & 2.5 & 8.5 & 8.4 & \B{2.6} & 8.4 & 2.7 & 2.3 & 2.4 & 2.8 & \B{2.5} & \B{2.5} \\
  5 & 1rotate1 & 6.7 & 6.9 & \BR{6.8} & 8.7 & 12.5 & 12.6 & \B{7.1} & 12.5 & 7.2 & 7.3 & 7.4 & 7.1 & \B{6.9} & \B{6.9} \\
  6 & 1rotate2 & 12.5 & 14.6 & \BR{12.6} & 14.8 & 18.5 & 18.4 & 13.6 & 18.2 & \B{13.4} & 14.2 & 14.3 & 13.6 & \B{13.2} & \B{13.2} \\
  7 & 1rotate3 & 13.0 & 24.3 & \BR{13.1} & 15.3 & 20.1 & 20.0 & \B{13.6} & 20.0 & 13.7 & 15.3 & 15.2 & 13.9 & \B{13.2} & \B{13.2} \\
  8 & 1rotate4 & 11.7 & 27.5 & \BR{11.8} & 13.3 & 18.8 & 18.6 & 12.4 & 18.6 & \B{12.1} & 13.7 & 13.4 & 12.8 & \B{12.1} & \B{12.1} \\
  9 & 1rotate5 & 11.0 & 25.2 & \BR{11.0} & 13.2 & 18.4 & 18.5 & \B{11.6} & 18.4 & \B{11.6} & 13.5 & 13.2 & 12.2 & \B{11.5} & 11.6 \\
  10 & 1rotate6 & 12.0 & 25.9 & \BR{12.1} & 14.0 & 19.3 & 19.1 & \B{12.7} & 19.1 & 12.9 & 14.2 & 14.4 & 13.3 & \BR{12.1} & 12.2 \\
  11 & 1rotate7 & 15.3 & 28.5 & \B{15.4} & 17.6 & 21.8 & 22.2 & \B{15.7} & 22.0 & 15.9 & 17.4 & 17.4 & 16.7 & \BR{15.3} & 15.4 \\
  12 & 1rotate8 & 23.4 & 33.8 & \BR{23.6} & 28.7 & 29.7 & 29.7 & 24.3 & 29.9 & \B{24.1} & 26.4 & 26.6 & 25.2 & \B{24.5} & \B{24.5} \\
  13 & 1rotate9 & 38.0 & \BR{38.3} & 38.4 & 39.4 & 42.5 & 42.5 & \B{38.9} & 42.5 & 39.0 & 39.9 & 39.8 & \B{39.2} & 39.7 & 39.7 \\
  14 & 1scale1 & 6.7 & 6.9 & \BR{6.8} & 7.5 & 12.5 & 12.6 & \B{7.1} & 12.6 & 7.2 & 7.4 & 7.4 & 7.1 & \B{6.9} & \B{6.9} \\
  15 & 1scale2 & 5.8 & 6.8 & \BR{5.9} & 6.5 & 11.8 & 11.7 & \B{6.4} & 11.6 & 6.5 & 7.0 & 6.6 & 6.4 & \B{6.1} & \B{6.1} \\
  16 & 1scale3 & 4.9 & 6.8 & \BR{5.0} & 5.7 & 10.9 & 10.7 & 5.6 & 10.7 & \B{5.5} & 6.3 & 5.8 & 5.5 & \B{5.1} & \B{5.1} \\
  17 & 1scale4 & 4.3 & 6.8 & \BR{4.2} & 4.9 & 10.3 & 10.3 & 4.8 & 10.3 & \B{4.7} & 5.7 & 5.2 & 4.9 & \B{4.5} & 4.6 \\
  18 & 1scale5 & 3.7 & 6.8 & \BR{3.8} & 4.5 & 9.9 & 9.9 & 4.2 & 9.9 & \B{4.0} & 5.4 & 4.8 & 4.5 & \B{4.0} & \B{4.0} \\
  19 & 1scale*1 & 6.6 & \BR{6.7} & 6.8 & 7.1 & 12.2 & 12.3 & 7.3 & 12.3 & \B{7.1} & 7.0 & 7.1 & 7.1 & \B{7.0} & \B{7.0} \\
  20 & 1scale*2 & 14.4 & 14.7 & \BR{14.6} & 15.2 & 19.8 & 19.8 & 15.4 & 19.8 & \B{15.3} & 15.6 & 15.4 & \B{14.9} & 15.2 & 15.2 \\
  21 & 1scale*3 & 17.0 & 19.0 & \BR{17.4} & 19.0 & 22.6 & 22.8 & 18.2 & 22.8 & \B{18.1} & 20.5 & 20.0 & 19.2 & \B{17.6} & 17.8 \\
  22 & 1scale*4 & 16.6 & 21.8 & \BR{17.0} & 18.3 & 22.2 & 22.2 & 17.8 & 22.2 & \B{17.7} & 22.0 & 21.0 & 19.7 & \B{17.3} & \B{17.3} \\
  23 & 1scale*5 & 15.2 & 24.3 & \BR{15.3} & 16.7 & 20.9 & 20.9 & 16.4 & 21.0 & \B{16.2} & 20.9 & 19.8 & 18.9 & \B{15.8} & \B{15.8} \\
  24 & 2dist1 & 20.7 & \BR{20.8} & \BR{20.8} & 21.3 & 21.9 & 22.0 & \B{21.1} & 22.0 & 21.4 & 21.6 & 21.5 & \B{21.0} & 21.2 & 21.2 \\
  25 & 2dist2 & 12.1 & \BR{12.1} & \BR{12.1} & 12.4 & 13.2 & 13.2 & \B{12.4} & 13.2 & \B{12.4} & 12.8 & 12.7 & \B{12.4} & 12.6 & 12.6 \\
  26 & 2dist3 & 5.2 & \BR{5.2} & \BR{5.2} & 5.5 & 6.5 & 6.5 & \B{5.4} & 6.5 & \B{5.4} & 5.9 & 5.7 & \B{5.4} & 5.7 & 5.7 \\
  27 & 2dist4 & 1.9 & \BR{1.9} & \BR{1.9} & 2.0 & 3.1 & 3.1 & \B{2.0} & 3.1 & \B{2.0} & 2.3 & 2.1 & \BR{1.9} & 2.0 & 2.0 \\
  28 & 2rotate1 & 5.2 & \BR{5.2} & \BR{5.2} & 5.9 & 6.5 & 6.5 & \B{5.4} & 6.5 & \B{5.4} & 5.9 & 5.7 & \B{5.4} & 5.7 & 5.7 \\
  29 & 2rotate2 & 8.7 & 9.6 & \BR{8.8} & 9.3 & 10.1 & 10.1 & \B{9.1} & 10.1 & \B{9.1} & 9.3 & 9.3 & \B{9.2} & \B{9.2} & \B{9.2} \\
  30 & 2rotate3 & 9.0 & 15.0 & \BR{9.1} & 9.6 & 10.6 & 10.6 & 9.4 & 10.6 & \B{9.3} & 9.6 & 9.5 & \B{9.4} & \B{9.4} & \B{9.4} \\
  31 & 2rotate4 & 8.0 & 21.3 & \BR{8.0} & 8.5 & 9.7 & 9.7 & \B{8.4} & 9.7 & \B{8.4} & 8.5 & 8.4 & \B{8.4} & \B{8.4} & \B{8.4} \\
  32 & 2rotate5 & 7.7 & 25.5 & \BR{7.7} & 8.1 & 9.3 & 9.4 & 8.0 & 9.4 & \B{7.9} & 8.1 & 8.0 & \B{8.1} & 8.2 & 8.2 \\
  33 & 2rotate6 & 8.3 & 25.5 & \BR{8.4} & 8.7 & 9.9 & 9.9 & \B{8.6} & 10.0 & \B{8.6} & 8.8 & 8.7 & \B{8.7} & \B{8.7} & 8.8 \\
  34 & 2rotate7 & 10.5 & 25.3 & \BR{10.5} & 11.1 & 12.1 & 12.1 & 10.9 & 12.1 & \B{10.8} & 11.1 & 11.0 & 10.8 & \B{10.7} & \B{10.7} \\
  35 & 2rotate8 & 16.2 & 24.6 & \B{16.2} & 17.1 & 17.6 & 17.6 & 16.5 & 17.6 & \B{16.4} & 16.9 & 16.8 & \BR{16.0} & 16.1 & 16.1 \\
  36 & 2rotate9 & 22.9 & \BR{22.9} & \BR{22.9} & 23.3 & 23.8 & 24.0 & \B{23.2} & 23.9 & \B{23.2} & 23.3 & 23.2 & 23.1 & \B{23.0} & \B{23.0} \\
  37 & 2scale1 & 5.2 & \BR{5.2} & \BR{5.2} & 5.5 & 6.5 & 6.6 & \B{5.4} & 6.5 & \B{5.4} & 5.9 & 5.7 & \B{5.4} & 5.7 & 5.7 \\
  38 & 2scale2 & 4.6 & 5.2 & \BR{4.7} & 4.9 & 6.0 & 6.0 & \B{4.8} & 5.9 & \B{4.8} & 4.9 & 4.8 & \B{4.8} & 5.0 & 5.0 \\
  39 & 2scale3 & 4.1 & 5.2 & \BR{4.1} & 4.2 & 5.4 & 5.4 & \B{4.2} & 5.4 & \B{4.2} & 4.1 & 4.2 & \B{4.3} & \B{4.3} & \B{4.3} \\
  40 & 2scale4 & 3.7 & 5.2 & \BR{3.7} & 3.8 & 5.1 & 5.0 & \B{3.8} & 5.0 & \B{3.8} & 3.8 & 3.9 & 3.9 & \B{3.8} & \B{3.8} \\
  41 & 2scale5 & 3.3 & 5.2 & \BR{3.4} & \BR{3.4} & 4.7 & 4.7 & 3.5 & 4.7 & \BR{3.4} & 3.5 & 3.6 & 3.6 & \BR{3.4} & 3.5 \\
  42 & 2scale*1 & 5.2 & \BR{5.2} & 5.3 & 5.5 & 6.5 & 6.5 & \B{5.5} & 6.5 & \B{5.5} & 5.8 & 5.6 & \B{5.4} & 5.6 & 5.6 \\
  43 & 2scale*2 & 15.0 & 15.9 & \BR{15.0} & 15.5 & 16.0 & 16.1 & \B{15.4} & 16.1 & \B{15.4} & 15.9 & 15.8 & \B{15.3} & 15.5 & 15.5 \\
  44 & 2scale*3 & 19.1 & 26.6 & \BR{19.1} & 19.5 & 20.2 & 20.4 & \B{19.4} & 20.4 & \B{19.4} & 20.8 & 20.8 & 20.2 & \B{19.3} & \B{19.3} \\
  45 & 2scale*4 & 18.8 & 27.2 & \BR{18.9} & 19.6 & 20.0 & 20.1 & \B{19.1} & 20.1 & 19.2 & 20.3 & 20.7 & 20.0 & \B{19.1} & \B{19.1} \\
  46 & 2scale*5 & 17.4 & 25.6 & \BR{17.5} & 17.9 & 18.8 & 18.7 & 17.9 & 18.8 & \B{17.8} & 19.1 & 19.2 & 18.7 & \B{17.7} & \B{17.7} \\
  47 & disks\_100x100 & 0.0 & 49.3 & 35.5 & \B{13.0} & 24.2 & 24.2 & 21.1 & 24.2 & \B{20.5} & 15.3 & 16.0 & 11.6 & \BR{11.3} & 13.0 \\
  48 & disks\_300x500 & 0.0 & 37.5 & 29.3 & \B{5.7} & 16.3 & 16.3 & 14.1 & 16.3 & \B{12.7} & 6.6 & 7.1 & 5.2 & \BR{5.0} & 5.2 \\
  49 & disks\_400x400 & 0.0 & 49.9 & 30.8 & \B{6.1} & 18.5 & 18.4 & 17.5 & 18.4 & \B{16.7} & 7.0 & 7.4 & 5.7 & \BR{5.5} & \BR{5.5} \\
  50 & disks\_80x120 & 0.0 & 36.7 & 26.2 & \B{11.8} & 24.1 & 24.1 & 16.5 & 24.1 & \B{16.4} & 14.8 & 15.8 & 10.9 & \BR{10.8} & 11.6 \\
   \hline
\end{tabular}
}
\end{table}

\begin{table}[h]
\centering
\caption{Error rates (in \%) of the pot-pot classifiers for simulated data sets. }
\smallskip
\label{t:errors_sim_potential}
\newcommand{\B}[1]{{\textbf{#1}}}
\newcommand{\R}[1]{{\red{#1}}}
\newcommand{\BR}[1]{{\red{\textbf{#1}}}}
\resizebox{\textwidth}{!}
{
\begin{tabular}{|Hrr|rrrrr|rrrrr|rrr|}
  \hline
 \multirow{2}{*}{H} & \multirow{2}{*}{dataset} & \multirow{2}{*}{Bayes} &
 \multicolumn{5}{c|}{Joint} & \multicolumn{5}{c|}{Separate} & \multicolumn{3}{c|}{Regressive separate}\\
 & & &
 diag. & $\alpha$ & \kNN & ROT & mM       & diag. & $\alpha$ & \kNN & ROT & mM    & diag. & $\alpha$ & \kNN\\
  \hline
1 & 1dist1 & 30.9 & \B{31.4} & 32.2 & 33.3 & 32.7 & \B{31.4} & \BR{31.2} & 31.8 & 33.2 & 32.3 & \BR{31.2} & \BR{31.2} & 31.8 & 33.2 \\
  2 & 1dist2 & 15.8 & \B{16.2} & 16.7 & 16.9 & 16.7 & \B{16.2} & \BR{16.0} & 16.3 & 16.5 & 16.4 & \BR{16.0} & \BR{16.0} & 16.6 & 16.5 \\
  3 & 1dist3 & 6.7 & \B{6.8} & 7.1 & 7.2 & 7.1 & \B{6.8} & \BR{6.7} & 6.9 & 7.0 & 7.0 & 6.8 & \B{6.8} & 6.9 & 7.1 \\
  4 & 1dist4 & 2.0 & \B{2.4} & 2.8 & 2.7 & 2.5 & 2.5 & \BR{2.3} & 2.5 & 2.5 & 2.4 & 2.4 & \BR{2.3} & 2.5 & 2.6 \\
  5 & 1rotate1 & 6.7 & \B{6.8} & 7.1 & 7.2 & 7.1 & \B{6.8} & \BR{6.7} & 6.9 & 7.0 & 7.0 & 6.8 & \B{6.8} & 6.9 & 7.1 \\
  6 & 1rotate2 & 12.5 & \B{13.4} & 13.6 & 14.3 & 13.5 & 14.4 & \BR{13.0} & 13.2 & 13.9 & 13.6 & \BR{13.0} & \BR{13.0} & 13.2 & 14.0 \\
  7 & 1rotate3 & 13.0 & \B{13.8} & 13.9 & 15.3 & 14.0 & 16.6 & \BR{13.1} & 13.2 & 13.8 & 13.4 & \BR{13.1} & \BR{13.1} & 13.2 & 13.8 \\
  8 & 1rotate4 & 11.7 & \B{12.5} & 12.8 & 14.1 & 12.8 & 15.3 & \BR{11.6} & 12.1 & 12.6 &11.9 & \BR{11.6} & \BR{11.6} & 12.1 & 12.7 \\
  9 & 1rotate5 & 11.0 & \B{12.2} & \B{12.2} & 13.7 & 12.3 & 14.8 & \BR{11.2} & 11.5 & 11.8 & 11.5 & \BR{11.2} & \BR{11.2} & 11.6 & 11.9 \\
  10 & 1rotate6 & 12.0 & \B{13.0} & 13.3 & 14.5 & 13.2 & 16.0 & \BR{11.9} & 12.1 & 12.8 & 12.3 & 12.0 & \BR{11.9} & 12.2 & 12.8 \\
  11 & 1rotate7 & 15.3 & \B{16.2} & 16.7 & 18.1 & \B{16.2} & 19.9 & \BR{15.0} & 15.3 & 16.0 & 15.3 & 15.1 & \BR{15.0} & 15.4 & 16.2 \\
  12 & 1rotate8 & 23.4 & \B{25.0} & 25.2 & 26.7 & \B{25.0} & 30.2 & \BR{23.8} & 24.5 & 25.4 & 24.3 & 23.9 & \BR{23.8} & 24.5 & 25.4 \\
  13 & 1rotate9 & 38.0 & \BR{38.6} & 39.2 & 40.8 & 40.3 & 38.7 & \B{39.5} & 39.7 & 42.1 & 40.7 & \B{39.5} & \B{39.5} & 39.7 & 42.1 \\
  14 & 1scale1 & 6.7 & \B{6.8} & 7.1 & 7.2 & 7.1 & \B{6.8} & \BR{6.7} & 6.9 & 7.0 & 7.0 & 6.8 & \B{6.8} & 6.9 & 7.1 \\
  15 & 1scale2 & 5.8 & \B{6.2} & 6.4 & 6.5 & \B{6.2} & 6.7 & \BR{6.0} & 6.1 & 6.1 & 6.4 & 6.2 & 6.2 & \B{6.1} & \B{6.1} \\
  16 & 1scale3 & 4.9 & \B{5.3} & 5.5 & 5.5 & 5.4 & 5.8 & \BR{5.0} & 5.1 & 5.1 & 5.4 & 5.2 & 5.6 & \B{5.1} & \B{5.1} \\
  17 & 1scale4 & 4.3 & \B{4.7} & 4.9 & 4.9 & 4.9 & 5.2 & \BR{4.2} & 4.5 & 4.4 & 4.8 & 4.6 & 5.4 & 4.6 & \B{4.4} \\
  18 & 1scale5 & 3.7 & \B{4.3} & 4.5 & 4.5 & 4.5 & 4.8 & \BR{3.8} & 4.0 & 4.1 & 4.4 & 4.0 & 5.1 & \B{4.0} & 4.2 \\
  19 & 1scale*1 & 6.6 & \B{6.9} & 7.1 & 7.2 & 7.0 & \B{6.9} & \BR{6.7} & 7.0 & 7.1 & 7.3 & \BR{6.7} & \BR{6.7} & 7.0 & 7.1 \\
  20 & 1scale*2 & 14.4 & \BR{14.9} & \BR{14.9} & 15.4 & \BR{14.9} & 15.0 & \B{15.0} & 15.2 & 15.3 & 16.0 & 15.3 & 16.4 & \B{15.2} & 15.3 \\
  21 & 1scale*3 & 17.0 & \B{19.0} & 19.2 & 20.1 & 19.1 & 19.5 & 17.8 & \BR{17.6} & 18.2 & 18.7 & 17.9 & \B{17.8} & \B{17.8} & 18.2 \\
  22 & 1scale*4 & 16.6 & \B{18.9} & 19.7 & 20.1 & 19.9 & 21.9 & \BR{17.3} & \BR{17.3} & 18.1 & 19.0 & 17.4 & 17.9 & \BR{17.3} & 18.1 \\
  23 & 1scale*5 & 15.2 & \B{17.9} & 18.9 & 18.7 & 19.1 & 21.6 & \BR{15.7} & 15.8 & 16.6 & 17.2 & 16.0 & 16.7 & \B{15.8} & 16.7 \\
  24 & 2dist1 & 20.7 & 21.1 & \BR{21.0} & 21.7 & 21.2 & \BR{21.0} & 21.4 & \B{21.2} & 21.7 & 21.5 & 21.3 & 21.5 & \B{21.2} & 21.7 \\
  25 & 2dist2 & 12.1 & \BR{12.4} & \BR{12.4} & 12.6 & 12.5 & 12.5 & \B{12.5} & 12.6 & 12.6 & \B{12.5} & 12.7 & \B{12.5} & 12.6 & 12.6 \\
  26 & 2dist3 & 5.2 & \BR{5.4} & \BR{5.4} & 5.6 & \BR{5.4} & \BR{5.4} & \B{5.6} & 5.7 & 5.7 & \B{5.6} & 5.7 & \B{5.6} & 5.7 & 5.7 \\
  27 & 2dist4 & 1.9 & \BR{1.9} & \BR{1.9} & 2.0 & 2.0 & 2.0 & \B{2.0} & \B{2.0} & \B{2.0} & 2.1 & 2.1 & \B{2.0} & \B{2.0} & 2.2 \\
  28 & 2rotate1 & 5.2 & \BR{5.4} & \BR{5.4} & 5.6 & \BR{5.4} & \BR{5.4} & \B{5.6} & 5.7 & 5.7 & \B{5.6} & 5.7 & \B{5.6} & 5.7 & 5.7 \\
  29 & 2rotate2 & 8.7 & \BR{9.0} & 9.2 & 9.3 &9.1 & 9.6 & \B{9.1} & 9.2 & 9.4 & \B{9.1} & 9.3 & \B{9.2} & \B{9.2} & 9.4 \\
  30 & 2rotate3 & 9.0 & \BR{9.2} & 9.4 & 9.7 & 9.4 & 12.0 & \B{9.3} & 9.4 & 9.5 & 9.4 & 9.5 & \B{9.4} & \B{9.4} & 9.6 \\
  31 & 2rotate4 & 8.0 & \BR{8.3} & 8.4 & 8.7 & 8.4 & 10.5 & \B{8.4} & \B{8.4} & 8.5 & \B{8.4} & \B{8.4} & \B{8.4} & \B{8.4} & 8.6 \\
  32 & 2rotate5 & 7.7 & \BR{8.0} & 8.1 & 8.4 & 8.2 & 10.1 & \B{8.1} & 8.2 & 8.2 & \B{8.1} & 8.2 & \B{8.1} & 8.2 & 8.2 \\
  33 & 2rotate6 & 8.3 & \BR{8.5} & 8.7 & 8.9 & 8.7 & 10.7 & \B{8.7} & \B{8.7} & 8.8 & \B{8.7} & 8.8 & \B{8.7} & 8.8 & 8.9 \\
  34 & 2rotate7 & 10.5 & \BR{10.6} & 10.8 & 11.0 & 10.8 & 13.4 & \B{10.7} & \B{10.7} & 10.9 & 10.8 & 10.8 & 10.8 & \B{10.7} & 10.9 \\
  35 & 2rotate8 & 16.2 & \BR{16.0} & \BR{16.0} & 16.5 & \BR{16.0} & 20.0 & 16.2 & \B{16.1} & 16.2 & 16.2 & 16.2 & 16.2 & \B{16.1} & 16.3 \\
  36 & 2rotate9 & 22.9 & \BR{23.0} & 23.1 & 23.2 & 23.2 & 23.1 & \B{22.9} & 23.0 & 23.1 & 23.2 & 23.1 & \BR{23.0} & \BR{23.0} & 23.1 \\
  37 & 2scale1 & 5.2 & \BR{5.4} & \BR{5.4} & 5.6 & \BR{5.4} & \BR{5.4} & \B{5.6} & 5.7 & 5.7 & \B{5.6} & 5.7 & \B{5.6} & 5.7 & 5.7 \\
  38 & 2scale2 & 4.6 & \BR{4.8} & \BR{4.8} & 5.0 & \BR{4.8} & 5.0 & 5.4 & \B{5.0} & 5.1 & 5.1 & 5.1 & 5.5 & \B{5.0} & 5.2 \\
  39 & 2scale3 & 4.1 & \BR{4.2} & 4.3 & 4.5 & 4.3 & 4.5 & 4.7 & \B{4.3} & 4.4 & 4.4 & \B{4.3} & 5.1 & \B{4.3} & 4.6 \\
  40 & 2scale4 & 3.7 & \B{3.9} & \B{3.9} & 4.2 & \B{3.9} & 4.4 & 4.1 & \BR{3.8} & \BR{3.8} & 3.9 & 3.9 & 4.3 & \BR{3.8} & 4.0 \\
  41 & 2scale5 & 3.3 & \B{3.6} & \B{3.6} & 3.9 & \B{3.6} & 4.3 & 3.8 & \BR{3.4} & 3.5 & 3.5 & 3.5 & 4.3 & \B{3.5} & 3.7 \\
  42 & 2scale*1 & 5.2 & 5.5 & \BR{5.4} & 5.6 & 5.5 & \BR{5.4} & \B{5.6} & \B{5.6} & 5.7 & \B{5.6} & 5.7 & 5.7 & \B{5.6} & 5.7 \\
  43 & 2scale*2 & 15.0 & 15.4 & \BR{15.3} & 15.8 & 15.4 & 15.4 & 17.7 & \B{15.5} & 15.7 & 15.8 & 15.7 & 19.0 & \B{15.5} & 15.7 \\
  44 & 2scale*3 & 19.1 & \B{20.2} & \B{20.2} & 20.6 & \B{20.2} & 21.0 & 20.4 & \BR{19.3} & 19.9 & 19.6 & \BR{19.3} & 21.6 & \BR{19.3} & 20.0 \\
  45 & 2scale*4 & 18.8 & 20.1 & \B{20.0} & 20.2 & 20.5 & 21.1 & 19.8 & \BR{19.1} & 19.5 & 19.3 & 19.2 & 20.6 & \BR{19.1} & 19.6 \\
  46 & 2scale*5 & 17.4 & \B{18.6} & 18.7 & 19.1 & 19.0 & 20.8 & 18.5 & \BR{17.7} & 18.1 & 18.0 & \BR{17.7} & 18.8 & \BR{17.7} & 18.1 \\
  47 & disks\_100x100 & 0.0 & 11.5 & 11.6 & \BR{7.8} & 22.9 & \BR{7.8} & 11.8 & 11.3 & \B{7.9} & 22.3 & 8.1 & 12.4 & 13.0 & \B{8.1} \\
  48 & disks\_300x500 & 0.0 & 5.6 & 5.2 & \BR{2.9} & 9.4 & \BR{2.9} & 4.8 & 5.0 & \B{3.4} & 9.3 & 3.7 & 5.0 & 5.2 & \B{3.5} \\
  49 & disks\_400x400 & 0.0 & 5.7 & 5.7 & \B{4.0} & 10.8 & \B{4.0} & 5.7 & 5.5 & \BR{3.6} & 11.8 & 3.8 & 5.8 & 5.5 & \B{3.8} \\
  50 & disks\_80x120 & 0.0 & 10.9 & 10.9 & \BR{6.7} & 18.6 & \BR{6.7} & 10.4 & 10.8 & \B{7.0} & 17.7 & 7.2 & 11.5 & 11.6 & \B{7.2} \\
   \hline
\end{tabular}
}
\end{table} 
\begin{table}[h]
\centering
\caption{Error rates (in \%) of different classifiers for real data sets.}
\parbox[t]{\textwidth}{Columns (7) -- (11) and (14) -- (16): $\alpha$-classifier in the $DD$- and pot-pot plots.}
\smallskip
\label{t:errors_depths}
\newcommand{\B}[1]{{\textbf{#1}}}
\newcommand{\R}[1]{{\red{#1}}}
\newcommand{\BR}[1]{{\red{\textbf{#1}}}}
\resizebox{1\textwidth}{!}
{
\begin{tabular}{|Hr|rr|rrr|rrrrr|rr|rrr|}
  \hline
  {H} & {(1)} & {(2)} & {(3)} & {(4)} &
 {(5)} & {(6)} & {(7)} & {(8)} & {(9)}         & (10)  & (11)
 & (12)  & (13)  & (14)   & (15)  & (16)\\
    {H} & {} & {} & {} & {} & {} & {} &
 {} & {} & {} & {} & {}         & Dknn  & Dknn
 & pot-pot  & pot-pot  & pot-pot \\
  {H} & {dataset} & {$N$} & $d$  & {LDA} & {QDA} & {KNN} &
 {Zonoid} & {HS} & {Mah} & {Proj} & {Spat}         & HS & Mah
 & joint & separate & regress.\\
    {H} & {} & {} & {} & {} & {} & {} &
 {} & {} & {} & {} & {}         &   &
 &  &  & separate \\
  \hline
1 & baby & 247 & 5 & 22.3 & 22.3 & \B{21.5} & 22.7 & 22.7 & 24.7 & \B{21.5} & 25.5 & 29.1 & 32.8 & \BR{20.2} & 23.1 & 24.3 \\
  2 & banknoten & 200 &	6 & \B{0.5} & \B{0.5} & \B{0.5} & \B{0.5} & \B{0.5} & 1.0 & \B{0.5} & 1.0 & 2.0 & 5.0 & 0.5 & \BR{0.0} & \BR{0.0} \\
  3 & biomed & 194 & 4 & 16.0 & \B{12.4} & \B{12.4} & 13.4 & \B{11.3} & 12.4 & 12.9 & 13.4 & 26.8 & 17.5 & 13.4 & \BR{9.3} & 9.8 \\
  4 & bloodtransfusion & 748 &	3 & 23.0 & 22.3 & \B{21.3} & 23.1 & 24.3 & \B{20.5} & 23.5 & 21.8 & 21.3 & 20.7 & 19.9 & \BR{19.0} & 19.7 \\
  5 & breast\_cancer\_wisconsin & 699 &	9 & 4.0 & 4.9 & \B{3.1} & 19.3 & 14.9 & \B{3.6} & 15.7 & \B{3.7} & 27.6 & 31.8 & 0.9 & \BR{0.7} & \BR{0.7} \\
  6 & bupa & 345 & 6 & \B{30.1} &  40.6 & 30.7 & 30.7 & 30.4 & 29.9 & \BR{27.8} & 29.9 & 30.7 & 31.3 & 29.9 & \B{29.0} & 29.9 \\
  7 & chemdiab\_1vs2 & 112 & 5 & \B{3.6} & 7.1 & 9.8 & \B{3.6} & \B{3.6} & \B{3.6} & \B{3.6} & \B{3.6} & 11.6 & 8.0 & \BR{1.8} & 2.4 & 3.3 \\
  8 & chemdiab\_1vs3 & 69 & 5 & 10.1 & \B{8.7} & \B{8.7} & 10.1 & 8.7 & 10.1 & \B{7.2} & \B{7.2} & 8.7 & 11.6 & 8.7 & \BR{5.8} & 7.2 \\
  9 & chemdiab\_2vs3 & 109 & 5 & 3.7 & \B{0.9} & \B{0.9} & 3.7 & 3.7 & \B{1.8} & 3.7 & \B{1.8} & 6.4 & 11.0 & 0.9 & \BR{0.0} & \BR{0.0} \\
  10 & cloud & 108 & 7 &  54.6 & \B{47.2} & 54.6 & 54.6 & 50.9 & \B{46.3} & 52.8 & 48.1 & 40.7 & 50.0 & 39.8 & \BR{32.4} & 39.8 \\
  11 & crabB\_MvsF & 200 & 5 & \B{9.0} & 10.0 & 14.0 & 9.0 & 8.0 & \B{6.0} & 8.0 & \B{6.0} & 9.0 & 16.0 & \BR{5.0} & \BR{5.0} & 6.0 \\
  12 & crabF\_BvsO & 200 & 5 & \BR{0.0} & 1.0 & 5.0 & \BR{0.0} & \BR{0.0} & 1.0 & \BR{0.0} & 1.0 & 2.0 & 1.0 & \BR{0.0} & \BR{0.0} & \BR{0.0} \\
  13 & crabM\_BvsO & 100 & 5 & \BR{0.0} & \BR{0.0} & 5.0 & \BR{0.0} & \BR{0.0} & \BR{0.0} & \BR{0.0} & \BR{0.0} & 2.0 & 3.0 & \BR{0.0} & \BR{0.0} & \BR{0.0} \\
  14 & crabO\_MvsF & 100 & 5 & 3.0 & \B{2.0} & 8.0 & 3.0 & 3.0 & \B{2.0} & 3.0 & \B{2.0} & 4.0 & 5.0 & 2.0 & \BR{1.0} & 2.0 \\
  15 & crab\_BvsO & 100 & 5 & \BR{0.0} & \BR{0.0} & 3.5 & \BR{0.0} & \BR{0.0} & \BR{0.0} & \BR{0.0} & \BR{0.0} & 0.5 & 1.0 & \BR{0.0} & \BR{0.0} & \BR{0.0} \\
  16 & crab\_MvsF & 100 & 5 & \B{4.0} & 5.0 & 9.0 & 4.0 & 3.5 & 4.5 & \BR{3.0} & 3.5 & 6.0 & 7.0 & 4.0 & \B{3.5} & 4.0 \\
  17 & cricket\_CvsP & 156 & 4 & 68.6 & 64.1 & \B{63.5} & \B{59.6} & 62.8 & 64.7 & 64.1 & 60.9 & 56.4 & 55.1 & 56.4 & \BR{47.4} & \BR{47.4} \\
  18 & diabetes & 768 &	8 & \B{22.4} & 26.6 & 25.1 & 33.9 & 30.1 & \B{24.6} & 29.4 & \B{24.7} & 28.6 & 29.3 & \BR{22.1} & 22.4 & 24.2 \\
  19 & ecoli\_cpvsim & 220 & 5 & \B{1.4} & 1.8 & 1.8 & \B{1.4} & 2.3 & \B{1.4} & 2.3 & \B{1.4} & 14.5 & 9.5 & \BR{0.9} & \BR{0.9} & 1.8 \\
  20 & ecoli\_cpvspp & 195 & 5 & \BR{3.1} & 4.1 & 3.6 & \B{4.1} & 4.6 & 4.6 & 4.6 & 5.1 & 14.4 & 9.2 & \BR{3.1} & 3.6 & 3.6 \\
  21 & ecoli\_imvspp & 129 & 5 & 5.4 & \B{3.9} & 6.2 & 5.4 & 5.4 & \B{2.3} & 5.4 & 3.9 & 4.7 & 6.2 & 2.3 & \BR{1.0} & \BR{1.1} \\
  22 & gemsen\_MvsF & 1349 & 6 & 19.1 & 14.2 & \B{14.1} & 14.8 & 16.4 & 14.8 & 16.5 & \B{14.0} & 10.8 & 12.5 & 10.9 & \BR{10.3} & 10.7 \\
  23 & glass & 146 & 9 & 27.4 & 39.7 & \B{18.5} & \B{27.4} & 30.8 & 30.1 & 33.6 & 28.1 & 33.6 & 30.8 & 23.4 & \BR{17.8} & 23.3 \\
  24 & groessen\_MvsF & 230 &	3 & 10.9 & \B{10.4} & 15.7 & \B{10.0} & 12.6 & 10.9 & 12.2 & 10.9 & 12.6 & 12.2 & 10.4 & \BR{7.8} & 9.6 \\
  25 & haberman & 306 &	3 & 25.2 & \B{24.5} & \B{24.5} & 26.8 & 28.1 & 27.1 & 26.5 & \B{25.5} & 28.8 & 29.7 & \BR{23.9} & \BR{23.9} & 25.2 \\
  26 & heart & 270 & 13 & \B{16.3} & 16.7 & 35.2 & \B{16.3} & 25.9 & 19.6 & 24.4 & 19.3 & 30.4 & 30.0 & 3.4 & \BR{0.0} & \BR{0.0} \\
  27 & hemophilia & 75 & 2 & \B{14.7} & 16.0 & 18.7 & 16.0 & \B{13.3} & 17.3 & \B{13.3} & 16.0 & 16.0 & 13.3 & \BR{12.0} & \BR{12.0} & \BR{12.0} \\
  28 & indian\_liver\_patient\_1vs2 & 579 & 10 & \B{29.7} & 44.6 & 32.0 & 29.4 & 30.4 & 30.7 & 30.1 & \B{28.5} & 28.2 & 28.3 & 27.9 & \BR{25.0} & 26.5 \\
  29 & indian\_liver\_patient\_FvsM & 579 & 9 & \B{24.5} & 63.0 & 25.4 & 24.9 & 24.9 & \B{24.7} & 26.1 & 25.7 & 24.7 & 24.2 & 24.4 & \BR{22.7} & \BR{22.7} \\
  30 & iris\_setosavsversicolor & 100 &	4 & \BR{0.0} & \BR{0.0} & \BR{0.0} & \BR{0.0} & \BR{0.0} & \BR{0.0} & \BR{0.0} & \BR{0.0} & \BR{0.0} & \BR{0.0} & \BR{0.0} & \BR{0.0} & \BR{0.0}  \\
  31 & iris\_setosavsvirginica & 100 &	4 & \BR{0.0} & \BR{0.0} & \BR{0.0} & \BR{0.0} & \BR{0.0} & \BR{0.0} & \BR{0.0} & \BR{0.0} & \BR{0.0} & \BR{0.0} & \BR{0.0} & \BR{0.0} & \BR{0.0} \\
  32 & iris\_versicolorvsvirginica & 100 & 4 & \B{3.0} & 4.0 & 6.0 & \B{3.0} & \B{3.0} & \B{3.0} & \B{3.0} & 6.0 & 3.0 & 7.0 & \BR{2.0} & \BR{2.0} & 3.0 \\
  33 & irish\_ed\_MvsF & 500 &	5 & 45.0 & \B{43.4} & 47.0 & 42.0 & \B{40.4} & 45.2 & 40.6 & 45.6 & 43.0 & 44.2 & 39.8 & \BR{37.4} & 37.6 \\
  34 & kidney & 76 & 5 & \B{28.9} & \B{28.9} & 32.9 & \B{28.9} & 30.3 & 30.3 & 30.3 & 31.6 & 30.3 & 38.2 & 23.7 & \BR{15.8} & 17.1 \\
  35 & pima & 200 &  7 & \B{24.5} & 27.5 & 29.5 & \B{25.5} & 27.5 & 28.5 & 26.0 & 30.0 & 30.5 & 35.5 & 25.0 & \BR{23.0} & 23.5 \\
  36 & plasma\_retinol\_MvsF & 315 & 13 & 14.3 & 14.0 & \B{13.7} & \B{14.3} & 15.9 & 14.6 & 17.1 & \B{14.3} & 13.7 & 12.7 & 12.0 & \BR{9.1} & 9.2 \\
  37 & segmentation & 660 & 10 & 8.2 & 9.4 & \B{4.5} & 6.8 & 6.1 & 9.2 & \B{4.8} & 8.8 & 5.9 & 4.2 & \BR{2.7} & \BR{2.7} & \BR{2.7} \\
  38 & socmob\_IvsNI & 1156 & 5 & 33.2 & 34.3 & \B{32.5} & \BR{27.9} & 33.6 & 31.6 & 33.3 & 30.4 & 35.3 & 46.6 & 33.6 & \B{28.8} & 29.5 \\
  39 & socmob\_WvsB & 1156 & 5 & 28.1 & 29.2 & \B{18.9} & \B{17.5} & 18.3 & 19.9 & 18.4 & 19.5 & 17.3 & 31.1 & 28.3 & \BR{16.4} & 16.7 \\
  40 & tae & 151 & 5 & \B{17.2} & 19.9 & 24.5 & \BR{11.9} & 13.2 & 17.2 & 17.2 & 17.2 & 25.8 & 18.5 & \B{13.2} & 13.9 & 14.6 \\
  41 & tennis\_MvsF & 87 & 15 & \B{41.4} & 44.8 & 43.7 & 41.4 & 44.8 & \B{36.8} & 46.0 & 36.8 & 48.3 & 46.0 & 31.2 & \BR{20.0} & \BR{20.0} \\
  42 & tips\_DvsN & 244 & 6 & 6.1 & \B{3.7} & 7.8 & 10.7 & 8.2 & \B{3.3} & 9.0 & 3.7 & 9.4 & 8.6 & 3.7 & \BR{2.9} & 3.3 \\
  43 & tips\_MvsF & 244 & 6 & 36.5 & 38.5 & \B{32.4} & 41.4 & 44.3 & 36.5 & 43.0 & \B{38.1} & 34.8 & 34.4 & 32.8 & \BR{28.7} & 32.4 \\
  44 & uscrime\_SvsN & 47 & 13 & 17.0 & 19.1 & \B{10.6} & \B{17.0} & \B{17.0} & 19.1 & \B{17.0} & 19.1 & 17.0 & 27.7 & \BR{0.0} & \BR{0.0} & \BR{0.0} \\
  45 & vertebral\_column & 310 & 6 & \B{15.5} & 17.4 & 16.5 & 15.8 & 15.8 & \B{14.5} & 17.4 & 15.8 & 15.8 & 16.5 & 15.2 & \BR{13.5} & 14.5 \\
  46 & veteran\_lung\_cancer & 137 &  7 & 64.2 & 51.8 & \B{50.4} & 62.0 & 57.7 & \B{47.4} & 57.7 & 50.4 & 48.9 & 48.2 & 40.1 & \BR{29.2} & 39.4 \\
  47 & vowel\_MvsF & 990 & 13 & 0.1 & 0.7 & \BR{0.0} & \B{0.1} & 24.0 & 0.4 & 24.8 & 0.5 & 0.3 & 0.2 & \BR{0.0} & \BR{0.0} & \BR{0.0} \\
  48 & wine\_1vs2 & 130 & 13 & \BR{0.0} & 0.8 & 5.4 & \BR{0.0} & 3.1 & 1.5 & 2.3 & 1.5 & 6.9 & 38.5 & \BR{0.0} & \BR{0.0} & \BR{0.0} \\
  49 & wine\_1vs3 & 107 & 13 & \BR{0.0} & \BR{0.0} & 13.1 & \BR{0.0} & 0.9 & \BR{0.0} & 1.9 & \BR{0.0} & 4.7 & 19.6 & \BR{0.0} & \BR{0.0} & \BR{0.0} \\
  50 & wine\_2vs3 & 119 & 13 & 0.8 & \BR{0.0} & 23.5 & 0.8 & 4.2 & \BR{0.0} & 4.2 & \BR{0.0} & 10.1 & 11.8 & \BR{0.0} & \BR{0.0} & \BR{0.0} \\
   \hline
\end{tabular}
}
\end{table}

\begin{table}[h]
\centering
\caption{Error rates (in \%) of the pot-pot classifiers for real data sets.}
\smallskip
\label{t:errors_potential}
\newcommand{\B}[1]{{\textbf{#1}}}
\newcommand{\R}[1]{{\red{#1}}}
\newcommand{\BR}[1]{{\red{\textbf{#1}}}}
\resizebox{\textwidth}{!}
{
\begin{tabular}{|Hr|rr|rrrrr|rrrrr|rrr|}
  \hline
 \multirow{2}{*}{H} & \multirow{2}{*}{dataset} & \multirow{2}{*}{$N$} & \multirow{2}{*}{$d$} &
 \multicolumn{5}{c|}{Joint} & \multicolumn{5}{c|}{Separate} & \multicolumn{3}{c|}{Regressive separate}\\
 & & & &
 diag. & $\alpha$ & \kNN & ROT & mM       & diag. & $\alpha$ & \kNN & ROT & mM    & diag. & $\alpha$ & \kNN\\
  \hline
1 & baby & 247 & 5 & 25.9 & \BR{20.2} & 24.3 & 31.6 & 21.5 & 24.3 & \B{23.1} & \B{23.1} & 33.2 & 23.9 & 25.1 & 24.3 & \B{23.1} \\
  2 & banknoten & 200 &	6 & 0.5 & 0.5 & \BR{0.0} & 2.0 & \BR{0.0} & \BR{0.0} & \BR{0.0} & \BR{0.0} & 1.0 & \BR{0.0} & \BR{0.0} & \BR{0.0} & \BR{0.0} \\
  3 & biomed & 194 & 4 & 14.4 & \B{13.4} & 14.9 & 16.5 & 14.4 & 10.8 & \BR{9.3} & 10.8 & 11.9 & 12.4 & 13.4 & \B{9.8} & 10.8 \\
  4 & bloodtransfusion & 748 &	3 & 21.3 & \B{19.9} & 20.5 & 20.7 & 21.4 & 20.1 & \BR{19.0} & 20.5 & 20.5 & 21.5 & 20.1 & \B{19.7} & 20.9 \\
  5 & breast\_cancer\_wisconsin & 699 &	9 & \B{0.9} & \B{0.9} & 3.6 & 4.1 & \B{0.9} & \BR{0.7} & \BR{0.7} & 3.0 & 8.0 & 2.5 & \BR{0.7} & \BR{0.7} & 3.9 \\
  6 & bupa & 345 & 6 &30.1 & \B{29.9} & 30.4 & 31.3 & 32.2 & 30.1 & \BR{29.0} & 30.1 & 31.9 & 31.0 & 32.8 & \B{29.9} & 30.1 \\
  7 & chemdiab\_1vs2 & 112 & 5 & 3.9 & \BR{1.8} & 2.7 & 8.0 & 3.9 & \B{2.4} & \B{2.4} & 3.6 & 7.1 & 3.3 & \B{2.5} & 3.3 & 4.5 \\
  8 & chemdiab\_1vs3 & 69 & 5 &  11.6 & \B{8.7} & \B{8.7} & 13.0 & \B{8.7} & 5.8 & 5.8 & \BR{4.3} & 13.0 & 7.2 & \B{7.2} & \B{7.2} & \B{7.2} \\
  9 & chemdiab\_2vs3 & 109 & 5 & 4.8 & \B{0.9} & 3.7 & 6.4 & \B{0.9} & \BR{0.0} & \BR{0.0} & \BR{0.0} & 8.3 & \BR{0.0} & \BR{0.0} & \BR{0.0} & \BR{0.0} \\
  10 & cloud & 108 & 7 &  40.7 & 39.8 & \BR{0.0} & 42.6 & \BR{0.0} & 38.9 & 32.4 & \BR{0.0} & 38.0 & \BR{0.0} & 38.9 & 39.8 & \B{17.1} \\
  11 & crabB\_MvsF & 200 & 5 & 10.0 & 5.0 & \BR{0.0} & 12.0 & \BR{0.0} & 6.0 & 5.0 & \BR{0.0} & 10.0 & \BR{0.0} & 9.0 & 6.0 & \BR{0.0} \\
  12 & crabF\_BvsO & 200 & 5 & \BR{0.0} & \BR{0.0} & \BR{0.0} & 2.0 & \BR{0.0} & \BR{0.0} & \BR{0.0} & \BR{0.0} & \BR{0.0} & \BR{0.0} & \BR{0.0} & \BR{0.0} & \BR{0.0} \\
  13 & crabM\_BvsO & 100 & 5 & \BR{0.0} & \BR{0.0} & \BR{0.0} & 2.0 & \BR{0.0} & \BR{0.0} & \BR{0.0} & \BR{0.0} & \BR{0.0} & \BR{0.0} & \BR{0.0} & \BR{0.0} & \BR{0.0} \\
  14 & crabO\_MvsF & 100 & 5 & 3.0 & 2.0 & \BR{0.0} & 4.0 & \BR{0.0} & 1.0 & 1.0 & \BR{0.0} & 3.0 & \BR{0.0} & 2.0 & 2.0 & \BR{0.0} \\
  15 & crab\_BvsO & 100 & 5 & 0.6 & \BR{0.0} & \BR{0.0} & 1.0 & \BR{0.0} & \BR{0.0} & \BR{0.0} & \BR{0.0} & \BR{0.0} & \BR{0.0} & \BR{0.0} & \BR{0.0} & \BR{0.0} \\
  16 & crab\_MvsF & 100 & 5 & 5.0 & 4.0 & \BR{0.0} & 6.5 & \BR{0.0} & 4.0 & 3.5 & \BR{0.0} & 6.5 & \BR{0.0} & 5.0 & 4.0 & \BR{0.0} \\
  17 & cricket\_CvsP & 156 & 4 & 67.9 & 56.4 & \BR{34.8} & 73.7 & \BR{34.8} & 48.1 & 47.4 & \BR{34.8} & 74.4 & \BR{34.8} & 48.7 & \B{47.4} & 54.5 \\
  18 & diabetes & 768 &	8 &  26.2 & \BR{22.1} & 23.0 & 27.6 & 22.7 & 24.3 & \B{22.4} & 23.8 & 28.5 & 24.9 & 25.5 & \B{24.2} & 24.7 \\
  19 & ecoli\_cpvsim & 220 & 5 &  2.7 & \BR{0.9} & 1.8 & 3.2 & \BR{0.9} & 1.4 & \BR{0.9} & \BR{0.9} & 2.7 & 2.3 & \B{1.8} & \B{1.8} & \B{1.8} \\
  20 & ecoli\_cpvspp & 195 & 5 & 3.6 & \BR{3.1} & 3.6 & 3.6 & 4.1 & \B{3.6} & \B{3.6} & \B{3.6} & 4.6 & 4.1 & \B{3.6} & \B{3.6} & \B{3.6} \\
  21 & ecoli\_imvspp & 129 & 5 & 2.9 & \B{2.3} & 3.1 & 3.9 & 2.9 & \BR{1.0} & \BR{1.0} & 3.1 & 3.1 & 1.1 & \B{1.1} & \B{1.1} & 3.1 \\
  22 & gemsen\_MvsF & 1349 & 6 & 10.6 & 10.9 & \B{10.5} & \B{10.5} & 13.3 & \BR{10.2} & 10.3 & 10.5 & \BR{10.2} & 13.4 & 11.0 & 10.7 & \B{10.5} \\
  23 & glass & 146 & 9 & 24.1 & 23.4 & \B{21.5} & 24.1 & 27.1 & 22.4 & \BR{17.8} & 25.3 & 26.9 & 27.8 & 26.2 & \B{23.3} & 25.3 \\
  24 & groessen\_MvsF & 230 &	3 & 10.0 & 10.4 & \B{9.6} & 10.9 & 10.4 & 8.3 & 7.8 & \BR{7.4} & 11.7 & 10.0 & \B{9.1} & 9.6 & 9.6 \\
  25 & haberman & 306 &	3 & 25.8 & \BR{23.9} & \BR{23.9} & 27.1 & 25.5 & 24.5 & \BR{23.9} & \BR{23.9} & 25.5 & 26.1 & 25.2 & 25.2 & \B{24.8} \\
  26 & heart & 270 & 13 & \B{3.4} & \B{3.4} & 17.4 & 23.7 & 5.3 & \BR{0.0} & \BR{0.0} & 12.5 & 23.7 & \BR{0.0} & \BR{0.0} & \BR{0.0} & 12.5 \\
  27 & hemophilia & 75 & 2 &  12.0 & 12.0 & \BR{9.3} & 12.0 & 14.7 & 13.3 & 12.0 & \BR{9.3} & 13.3 & 16.0 & 13.3 & 12.0 & \B{10.7} \\
  28 & indian\_liver\_patient\_1vs2 & 579 & 10 & 28.4 & 27.9 & \B{27.8} & 29.1 & 28.5 & 26.0 & \BR{25.0} & 25.9 & 26.8 & 28.5 & \B{26.3} & 26.5 & \B{26.3} \\
  29 & indian\_liver\_patient\_FvsM & 579 & 9 &  \B{24.2} & 24.4 & \B{24.2} & 24.4 & \B{24.2} & \BR{22.7} & \BR{22.7} & 22.8 & 24.3 & 24.0 & \BR{22.7} & \BR{22.7} & 23.5 \\
  30 & iris\_setosavsversicolor & 100 &	4 & \BR{0.0} & \BR{0.0} & \BR{0.0} & \BR{0.0} & \BR{0.0} & \BR{0.0} & \BR{0.0} & \BR{0.0} & \BR{0.0} & \BR{0.0} & \BR{0.0} & \BR{0.0} & \BR{0.0} \\
  31 & iris\_setosavsvirginica & 100 &	4 & \BR{0.0} & \BR{0.0} & \BR{0.0} & \BR{0.0} & \BR{0.0} & \BR{0.0} & \BR{0.0} & \BR{0.0} & \BR{0.0} & \BR{0.0} & \BR{0.0} & \BR{0.0} & \BR{0.0} \\
  32 & iris\_versicolorvsvirginica & 100 & 4 & 2.0 & 2.0 & \BR{0.0} & 4.0 & \BR{0.0} & 3.8 & 2.0 & \BR{0.0} & 5.0 & \BR{0.0} & 3.8 & 3.0 & \BR{0.0} \\
  33 & irish\_ed\_MvsF & 500 &	5 & 42.8 & 39.8 & \B{38.5} & 42.6 & \B{38.5} & 39.2 & \BR{37.4} & 38.4 & 40.0 & 40.0 & 41.0 & \B{37.6} & 39.0 \\
  34 & kidney & 76 & 5 &  26.0 & 23.7 & \B{14.9} & 23.7 & \B{14.9} & \BR{13.2} & 15.8 & 13.8 & 25.0 & 13.8 & \BR{13.2} & 17.1 & 13.8 \\
  35 & pima & 200 &  7 & 25.5 & 25.0 & \BR{20.7} & 28.5 & \BR{20.7} & 25.5 & 23.0 & \B{21.4} & 30.5 & 22.0 & 25.5 & 23.5 & \B{21.4} \\
  36 & plasma\_retinol\_MvsF & 315 & 13 & 12.0 & 12.0 & \BR{8.3} & 16.2 & 12.7 & \B{9.1} & \B{9.1} & \B{9.1} & 14.3 & 13.3 & \B{9.2} & \B{9.2} & \B{9.2} \\
  37 & segmentation & 660 & 10 & \BR{2.7} & \BR{2.7} & 4.5 & 5.0 & \BR{2.7} & \BR{2.7} & \BR{2.7} & 5.0 & 5.2 & 2.8 & \BR{2.7} & \BR{2.7} & 5.0 \\
  38 & socmob\_IvsNI & 1156 & 5 & 40.3 & 33.6 & \B{32.4} & 45.9 & 33.7 & 31.1 & \BR{28.8} & 30.6 & 30.4 & 31.7 & 31.2 & \B{29.5} & 30.6 \\
  39 & socmob\_WvsB & 1156 & 5 & 30.5 & \B{28.3} & 28.4 & 29.5 & 28.7 & 17.1 & \BR{16.4} & 17.0 & 17.0 & 20.8 & 17.7 & \B{16.7} & 17.0 \\
  40 & tae & 151 & 5 & 16.6 & \B{13.2} & 13.9 & 15.9 & 14.4 & 14.6 & 13.9 & \BR{12.6} & 15.2 & 14.5 & 14.6 & 14.6 & \B{13.2} \\
  41 & tennis\_MvsF & 87 & 15 & \B{31.2} & \B{31.2} & 32.2 & 36.8 & \B{31.2} & \BR{20.0} & \BR{20.0} & 26.3 & 34.5 & 30.0 & \BR{20.0} & \BR{20.0} & 26.3 \\
  42 & tips\_DvsN & 244 & 6 & 4.1 & \B{3.7} & 4.5 & 6.6 & \B{3.7} & 3.7 & \BR{2.9} & \BR{2.9} & 5.3 & 3.3 & 4.1 & \B{3.3} & 4.5 \\
  43 & tips\_MvsF & 244 & 6 & 35.7 & \B{32.8} & 33.8 & 43.0 & 33.8 & 33.2 & 28.7 & \BR{32.0} & 39.8 & 33.6 & 33.6 & \B{32.4} & 32.8 \\
  44 & uscrime\_SvsN & 47 & 13 & \BR{0.0} & \BR{0.0} & \BR{0.0} & 17.0 & 12.8 & \BR{0.0} & \BR{0.0} & \BR{0.0} & 19.1 & 17.0 & \BR{0.0} & \BR{0.0} & \BR{0.0} \\
  45 & vertebral\_column & 10 & 6 & 17.4 & \B{15.2} & 16.5 & 17.4 & \B{15.2} & 16.5 & \BR{13.5} & 14.2 & 17.1 & 15.2 & 19.7 & 14.5 & \B{14.2} \\
  46 & veteran\_lung\_cancer & 37 &  7 & 43.0 & \B{40.1} & 40.9 & 43.1 & 44.6 & 42.1 & \BR{29.2} & 33.6 & 43.1 & 44.5 & 42.4 & 39.4 & \B{35.0} \\
  47 & vowel\_MvsF & 990 & 13 & \BR{0.0} & \BR{0.0} & 0.1 & 0.1 & \BR{0.0} & \BR{0.0} & \BR{0.0} & 0.1 & 0.2 & \BR{0.0} & \BR{0.0} & \BR{0.0} & 0.1 \\
  48 & wine\_1vs2 & 130 & 13 & \BR{0.0} & \BR{0.0} & 0.8 & 6.9 & \BR{0.0} & \BR{0.0} & \BR{0.0} & \BR{0.0} & \BR{0.0} & 1.5 & \BR{0.0} & \BR{0.0} & \BR{0.0} \\
  49 & wine\_1vs3 & 107 & 13 & \BR{0.0} & \BR{0.0} & \BR{0.0} & 2.8 & \BR{0.0} & \BR{0.0} & \BR{0.0} & \BR{0.0} & \BR{0.0} & \BR{0.0} & \BR{0.0} & \BR{0.0} & \BR{0.0} \\
  50 & wine\_2vs3 & 119 & 13 & \BR{0.0} & \BR{0.0} & 1.7 & 5.0 & 1.7 & \BR{0.0} & \BR{0.0} & \BR{0.0} & 1.7 & \BR{0.0} & \BR{0.0} & \BR{0.0} & \BR{0.0} \\
   \hline
\end{tabular}
}
\end{table}

\end{document}